\documentclass[10pt,journal,compsoc]{IEEEtran}
\usepackage{graphicx}
\usepackage{amsmath,amssymb} 
\usepackage{color}
\usepackage{soul}
\usepackage{amsfonts}
\usepackage[utf8]{inputenc}
\usepackage{algpseudocode}
\usepackage{blindtext}
\usepackage{graphicx}
\usepackage{tikz}
\usepackage{makecell}
\usepackage{subcaption}
\usepackage{lineno}
\usepackage{multirow}
\usepackage{fancyhdr}
\usepackage{amsthm}
\usepackage{textcomp}
\usepackage{amsmath}
\usepackage{mdwmath}
\usepackage{bm}
\usepackage{adjustbox}
\usepackage{lineno,hyperref}
\usepackage{physics}
\usepackage{amsmath}
\modulolinenumbers[5]
\usepackage{commath}
\usepackage{algorithm}
\usepackage{bm}
\usepackage{float}
\usepackage{soul}

\usepackage[sorting=none]{biblatex}
\begin{document}
\title{Responsible AI: Gender bias assessment in emotion recognition}

\author{Artem Domnich 
        and Gholamreza~Anbarjafari,~\IEEEmembership{Senior~Member,~IEEE} 

\IEEEcompsocitemizethanks{\IEEEcompsocthanksitem A. Domnich is with Microsoft Inc., Tallinn, Estonia.\protect\\
\IEEEcompsocthanksitem G. Anbarjafari is with PwC Advisory, Helsinki, Finland and iVCV O\"U, Tartu 51011, Estonia.
\protect\\
E-mail: shahab.anbarjafari@pwc.com}
\thanks{}
}

\markboth{}
{Domnich and Anbarjafari: Gender bias assessment in emotion recognition }

\IEEEtitleabstractindextext{%
\begin{abstract}
Rapid development of artificial intelligence (AI) systems amplify many concerns in society. These AI algorithms inherit different biases from humans due to mysterious operational flow and because of that it is becoming adverse in usage. As a result, researchers have started to address the issue by investigating deeper in the direction towards Responsible and Explainable AI. Among variety of applications of AI, facial expression recognition might not be the most important one, yet is considered as a valuable part of human-AI interaction. Evolution of facial expression recognition from the feature based methods to deep learning drastically improve quality of such algorithms. This research work aims to study a gender bias in deep learning methods for facial expression recognition by investigating six distinct neural networks, training them, and further analysed on the presence of bias, according to the three definition of fairness. The main outcomes show which models are gender biased, which are not and how gender of subject affects its emotion recognition. More biased neural networks show bigger accuracy gap in emotion recognition between male and female test sets. Furthermore, this trend keeps for true positive and false positive rates. In addition, due to the nature of the research, we can observe which types of emotions are better classified for men and which for women. Since the topic of biases in facial expression recognition is not well studied, a spectrum of continuation of this research is truly extensive, and may comprise detail analysis of state-of-the-art methods, as well as targeting other biases.
\end{abstract}

}

\maketitle

\IEEEraisesectionheading{\section{Introduction}\label{sec:introduction}}
\IEEEPARstart{R}{ising} awareness regarding Artificial Intelligence (AI) solutions brings up questions in social sectors, such as influence on job market, compliance with laws and ethics \cite{smith2014ai, neri2020artificial, bonnefon2016social, zou2018ai, AI_awer}. Human Computer Interaction (HCI) reaches unbelievable level of entanglement between computers and mankind. Although AI is only a set of mathematical approaches (a lot of them were discovered 50 years ago, like perceptron \cite{rosenblatt1958perceptron}, backpropagation \cite{ linnainmaa1970representation} or neural network \cite{qian1988predicting}), with rapidly grow of computational power, these methods have started to be adopted in various scientific and applied applications and thus AI has become flagships of science fleet.

Solutions, development directions, daily affairs and sometimes even human fates became completely depended on sophisticated AI algorithms which are acting as black boxes. Unfair credit scoring \cite{AI_aff1}, car crash and even death caused by autopilot system \cite{AI_aff2, AI_aff3}, racist profilisation \cite{AI_aff4} and unexpected behaviour of AI defence robots \cite{AI_aff5} are nowadays reality. However, while scientists together with industries develop and integrate this kind of solutions into the real world application, customers (and other researchers) start to examine outcomes, questioning and  dispute machine's will \cite{kingston2018artificial, carneiro2014online, stephens2020recent}. 

Since social concerns have been growing, more frequently collocation such as Responsible AI (RAI) \cite{Rise_RAI} or Explainable AI (XAI) \cite{Rise_XAI} appears on the horizon of the research world. These concerns had reached governmental levels, reinforced by law and decrees, they became a serious issues, which scientific society and companies started to tackle. As could be derived from definitions, RAI framework comprises steps and actions for utilization of AI in responsible way. Meanwhile, XAI is about how to justify, explain and prove the background, behaviour and the decisions of algorithms.

From these perspectives, all methods, which directly utilize personal data, came under close supervision of public. Nowadays, face detection and face recognition technologies have become ubiquitous and considered as something casual and simple. However, there are plenty other applications where the main input is a human face, although objectives might extremely vary. Digging deep in the jungle of AI technologies, this research work investigates on facial expression recognition (FER) in the center of RAI in order to facilitate the future research studies in this field.

One of the pillars of RAI is fairness. In simple words, AI algorithms should not be biased towards one or another group of people. Moreover, this convention could be rephrased as awareness of presence of bias by mentioning it, rather than hiding it. Since supervised machine learning algorithms are completely rely on training data, they often unintentionally inherit latent biases. The goal of this research is to investigate presence or absence of gender bias in deep learning approaches for FER. Usually, to claim whether a model is biased or not, researchers have to perform the same experiment on distinct test groups, which are sampled by target feature, which in this case is gender: male or female.

To conduct a research, several type of deep learning algorithms were chosen. These are neural networks either mixed type (convolutional recurrent neural network) or 3D convolutional neural networks (3D-CNN). Furthermore, training of particular neural networks were performed in two different ways: from the scratch and with pre-trained weights. The main source of data was a SASE-FE database, as described in \ref{sec:sase-fe}, which is provided by the iCV lab at University of Tartu. 

The contribution of this work is on fairness assessment in the context of RAI by reporting capabilities of different deep learning models for FER in terms of gender bias. In this paper, it has been denoted several definitions of fairness and according to them found which architectures are more gender biased and which are less. It also has been discovered, which emotions are more exaggerated and better recognized depends on gender.

\section{RELATED WORK}
Now we are reviewing the general information regarding responsible usage of AI, understanding facial expressions, and overview of deep learning method for FER. Furthermore, this section describes definition of biases, principles of fairness and has overview of related study about latter in machine learning. 
    
Spreading of AI technologies over various areas of mankind presence is allowing to achieve incredible performance and fascinating results. Computer vision, which takes honorable place in AI adopted fields, became ubiquitous used by different industries. Nowadays, computer vision applications became pivoting technologies in autonomous driving systems \cite{bojarski2016end, tammvee2020human}, video surveillance \cite{ko2018deep}, virtual and augmented reality \cite{marchand2015pose, kaminska2020stress}, robotics \cite{loncomilla2016object, bolotnikova2017real, bolotnikova2017circuit} and others \cite{kayalibay2017cnn, gomez2015multimodal, kamilaris2018deep, guo2018dominant, avots2021ensemble, Aktas2021}. This increases the immediate need of investigation on ethical and moral rules in order to assure that such assistive systems will not interfere with human rights.
    
Human emotion is an extremely wide field of research. During last decades, people were trying to justify, classify and investigate emotions \cite{niedenthal2012social, reiman1997neuroanatomical, noroozi2018survey, anbarjafari2018book, dolan2002emotion}. Emotion recognition, as a sub-domain of research, tries to find an answer on how to classify emotion which is displaying by human being. Overall, we, as humans, rely on different modalities, which are taken into the account each time when we try to recognize an emotion. Usually, the indicators are facial expression, articulation, gesticulation, body pose and context of the situation. Worth noticing, that emotion recognition is deeply entangled with psychology\cite{carlson1992psychology}.
    
However, still one of the easiest way to express and to understand an emotion is a facial expression. Even though, the emotion in facial expression (that is being displayed by a person) with high certainty can be distinguished even without context and additional information. Hence, over the time, FER became an independent task in computer vision domain, usually interpreted as emotion recognition. As a result, this is a task on the edge of psychology and computer vision, where from the first perspective scientists try to determine an emotion based only on face expression and from the second one, to study algorithms to classify them. There are various domains in which application of FER have been utilized \cite{liu2014deeply, chen2015augmented, kolakowska2014emotion}.
     
Usually emotions are described through Action Units (AU) \cite{tian2001recognizing} or determined as a point in Valence-Arousal (V-A) space \cite{russell1980circumplex}. The AU methodology does not determine emotion explicitly, but provides a set of state for each different parts of the face, such as "lips apart", "outer brow raiser" and so on. Scientists can denote a displayed emotion as a group of AUs. Opposite from AUs, V-A space represents human emotion from psychological perspective, i.e. group them on imaginary Cartesian space in such way that the distance between similar emotions is lower than the opposite ones.
    
Regardless of notation, scientists determine six basic human emotions: anger, disgust, fear, happiness, sadness and surprise \cite{ekman1999basic}. More complex, compound emotions, are usually contains two basic emotion and arbitrary number of additional ones \cite{loob2017dominant, guo2018dominant}.
    
Data for FER usually presented as still images, sequence of images or videos on which person is displaying emotion. Usually, depends on purpose of data, there are artificial (laboratory) \cite{sapinski2018multimodal} and real-world (wild) \cite{huang2008labeled, dhall2015video} type of data. The first type of data is usually collected in laboratories under special circumstances with particular requirements. The latter one is often collected from the internet, gathering images with an open access and label them.
        
Before discovering that deep machine learning algorithms has huge efficiency in image domain, approaches for FER typically were organized as a three stage pipeline: image preprocessing, feature extraction and classification. Feature extraction is a core of pipeline and these approaches were mainly distinguishable by method of feature extraction. 
    
The most famous and common methods are Gabor Feature Extraction and local binary pattern (LBP) \cite{ahonen2004face, feng2007facial}. The first method uses a set of various Gabor filters \cite{mehrotra1992gabor}, named Gabor Filter Bank, to extract particular features from image. This method is robust against scale, rotation and luminous intensity. \cite{lyons1998coding, zhang2012facial, mattela2018facial} are good examples of usage of Gabor filters for encoding facial expressions. The second approach is based on the histograms of binary maps, which are local representations of relation between target pixel and its neighbours: if the center pixel of the map is greater than a neighbour, a corresponding map value equals to 0, otherwise - 1. This simple yet effective approach has found extension in several works \cite{jabid2010robust, wang2012facial, chao2015facial}, consequently improving a technique. 
    
In terms of classifiers, the choices are well known and proven machine learning algorithms. Essential methods such as k-Nearest Neighbours (kNN) \cite{sohail2007classification}, Naive Bayes \cite{mao2016hierarchical}, Space Representation-based Classifier (SRC) \cite{huang2010new} and Support Vector Machine (SVM) \cite{tsai2018facial} where used intensively.
    
In spite of the tremendous works that has been done in development of algorithms for FER, after resounding success of AlexNet \cite{krizhevsky2012imagenet} in 2012, it was only matter of time, when deep learning approaches would take a main stage in FER. In 2015, \cite{chao2015facial} and \cite{burkert2015dexpression} presented convolutional neural networks (CNN) for FER, trained on LBP and still images correspondingly. Although feature-absence method showed much worse results, ability of deep learning algorithms to extract features independently had been already well known, therefore, research is this direction was continued. The variety of works in the following years have shown extreme performance gain, with respect to to the classical approaches \cite{zhao2016peak, chen2017convolution, yang2017facial, fathallah2017facial}. 
    
Most recent CNN approaches have gone far beyond and involve sophisticated ways to tackle a very specific problems which rise only in FER domain. For example, Pyramid With Super Resolution \cite{vo2020pyramid} shows extreme performance of FERPlus dataset \cite{barsoum2016training}. Facial Motion Prior Networks \cite{chen2019facial} take lead position in AffectNet \cite{mollahosseini2017affectnet} benchmark. One of the fastest network is MicroExpNet \cite{cugu2019microexpnet}, which also shows incredible performance on Oulu-CASIA dataset \cite{zhao2011facial}. 
    
    
Success in still image domain naturally was desired to be transferred onto domain of sequence of images or videos. The two main ways how to address sequential data with deep learning are recurrent based methods and 3D convolution based methods \cite{huang2019facial}. The target for both is to utilize temporal relation between frames to catch a displayed emotion over time. However, most of the recurrent neural networks are using CNN as a feature extractor, meanwhile 3D CNNs are trained from the scratch and usually do not utilize any additional networks.
    
There are many different ways of how union of recurrent neural network (RNN) and CNN feature extractor have been utilized to achieve necessary goals. In \cite{jaiswal2016deep} using bi-directional Long-Short Term Memory (LSTM) and CNN, a compound model has been trained for AUs recognition. As a competitors in EmotiW2016 \cite{dhall2016emotiw} challenge, authors of \cite{fan2016video} used both CNN-RNN and 3D CNN models to extract necessary features to train a model. This kind of approach requires a lot of computational power, yet not showing state-of-the-art performance, therefore was not widely used in future. \cite{kim2017multi} is another good example of usage a combination of CNN and LSTM. CNN is trained using frames of different intensity expression-states, meanwhile LSTM utilizes features which are extracted with aforementioned CNN to learn temporal representations. 

In \cite{yu2018spatio} authors designed a novel architecture which comprised 3D CNN and Nested LSTM. In detail, each output feature map of 3D convolution layers is passed to the Multi-dimensional Spatial Pyramid Pooling (extension of Spatial Pyramid Pooling \cite{he2015spatial}), forming a feature vector of fixed length. These vectors are served as an input to the first so called T-LSTM, modeling temporal relations. Next, hidden states of T-LSTM are going to C-LSTM, learning dependencies between convolutions. As a results, according to the experiments, this method had outperformed state-of-the-arts methods of that time.
    
In the area of 3DCNNs, the types of proposed methods were not that diverse, due to the extremely huge consumption of computational capacity. In 2014, \cite{byeon2014facial} authors proposed a simple architecture, showing that this method outperform existed classic approached and heavily depend on the size of the network. Three years later, \cite{hasani2017facial} presented a method, which comprises 3D convolutional neural network together with facial landmarks. Authors enhanced Inception-ResNet architecture \cite{szegedy2017inception} within 3D convolution and decreased model size in favor of compactness. As a results, this model beats all that time state-of-the-art methods. Recent years, this kind of model has found appliance in the micro expression recognition \cite{zhi2019combining, li2019micro, wu2021tsnn}.

    
Although deep learning has opened new horizons for many fields, thoughtless usage of this technology may lead to undesirable consequence. Recent years many problems have arose due to implementation of AI algorithm in fields with critical importance. In 2019, high-level expert group on AI released Ethics Guidelines For Trustworthy AI \cite{Trush_AI}. This document defines what trustworthy AI is and determine a framework which has to help companies adjust their AI technologies. According to the framework, three main components of trustworthy AI are lawful, robust and ethical.
    
Nowadays, a lot of companies invest in research and utilize these principles. For example, Microsoft has own Microsoft AI Principles\cite{Microsoft_AI} which are correlated with aforementioned guidelines. Corporation tries to bring responsible AI through the Office of Responsible AI (ORA) and the AI, Ethics and Effects in Engineering and Research (Aether) Committee. 
    
According to the PwC 2019 AI predictions\cite{PwC_AI_19}, only 3\% of companies do not have plans to address problems of RAI. This is a solid proof that rising concerns regarding AI are pushing industries to make AI more explainable and responsible. However, following 2020 AI predictions\cite{PwC_AI_20}, only one third of surveyed companies address problems of AI Ethics.
    
According to Ethics guidelines for trustworthy AI \cite{Trush_AI}, the one of the Ethical Principles in the Context of AI system is the principle of fairness. Fairness is a one out of four main principles of AI Ethics, along with respect for human autonomy, prevention of harm and explicability. However, this term is not unequivocally determined, opening a lot of doors to enter a discussion \cite{saxena2019fairness}. Bias in machine learning could be represented from the very different perspectives: biased sampling, inappropriate feature selection, inference of "black box" models, biased labeling and others. As has been mentioned before, AI industry is currently tackling risks of AI, in particular, making efforts to find and mitigate different kind of biases in machine learning.
    
Despite broad discussion, wide range of options and different definitions, in scope of this paper, fairness is treated in three different ways:
    \begin{enumerate}
        \item \textit{Equalized odds.} Predictor $\hat{Y}$ is fair, if for protected attribute $A$ and prediction $Y$, $\hat{Y}$ and $A$ are independent conditional on $Y$\cite{hardt2016equality}. In math formula
        \begin{equation}
        \label{form:eqod}
            P(\hat{Y} | A=0, Y=y) = P (\hat{Y} | A=1, Y=y)
        \end{equation}
        
        \item \textit{Equal Opportunity} \cite{hardt2016equality} defines this type of fairness for binary predictor as follow:
        \begin{equation}
        \label{form:eqop}
            P(\hat{Y} | A=0, Y=1) = P(\hat{Y} | A=1, Y=1)
        \end{equation}
        
        \item \textit{Demographic Parity.} Predictor $\hat{Y}$ is fair, if for protected attribute A and prediction $Y$ the likelihood of prediction is the same\cite{kusner2017counterfactual}. In math formula
        \begin{equation}
        \label{form:dp}
            P(\hat{Y}|A=0) = P (\hat{Y}|A=1)
        \end{equation}

    \end{enumerate}
    
Direct discrimination \cite{mehrabi2019survey} assumes biased results which are depended on particular attributes. Hence, usage of these attributes are considered as protected and defended by law \cite{chen2019fairness, peters2020responsible, cheng2021socially}.
    
Explainable and Non-explainable discrimination are two sides of one coin. Explainable discrimination is legal in situation, when particular biases could be explained and justified. Meanwhile, non-explainable discrimination is illegal, and being considered such, when discrimination toward particular groups could not be explained. However, special techniques are existed to identify and even mitigate illegal discrimination \cite{kamiran2013explainable}.
    
Since deep learning does not have clear features as input and decision are not explainable, often, due to the nature of data, different biases are arisen. There are more than 20 different biases \cite{mehrabi2019survey}. For example, Representation Bias \cite{suresh2019framework} touches most of well known datasets (ImageNet \cite{deng2009imagenet}, OpenImages \cite{krasin2017openimages}), since the population of data is biased towards one or another demographic group \cite{shankar2017no}. Another common example is Population Bias, which "arises when statistics, demographics, representatives, and user characteristics are different in the user population represented in the dataset or platform from the original target population" \cite{olteanu2019social}. In other words, Representation bias is hidden inside the way of gathering data from the population, meanwhile Population Bias might be stashed in the population itself.
    
Last couple of years, scientists around the world are trying to find out whether publicly available AI systems are biased toward protected attributes such as age, gender and race. And indeed, many of them are. Usually, AI is represented as trained neural networks, therefore, mimicking human behavior and being trained on datasets gathered by humans, these networks overtake one of the most prevailing bias - race bias. However, racist behaviour by default could not be expressed by AI, since it does not have own mind. Hence, usually race bias is inherited implicitly from the data, on which neural network has been trained. An excellent examples are researches about a race bias in face recognition \cite{klare2012face, el2016face, cook2019demographic}.
    
Although there are many existed biases, in this paper, we focus on a gender bias in FER. Unfortunately, only few studies have been conducted on examining different biases in FER, particularly gender bias. In \cite{howard2017addressing} authors target underrepresented class, showing discrimination in AI for FER solutions. Race bias has been found in Microsoft's Face and Face++ API's by \cite{rhue2018racial}. Several studies have proven\cite{denton2019detecting, wang2020towards} a correlation between gender and smile on the face (which implicitly mean positive emotion) in CelebA dataset \cite{liu2015deep}. \cite{xu2020investigating}, one of the most recent complete studies of bias and fairness in FER targets three different biases: age, race and gender. Authors of this research perform experiments on aforementioned CelebA \cite{liu2015deep} database and RAF-DB \cite{li2017reliable}. However, although authors divide test data into entirely male and female groups, they do not provide conclusions regarding per class differences in accuracy. Instead of that, they show that model is biased towards female group, gaining overall 3\% less accuracy score. Another outcome from the study is that according to the \textit{Equal opportunity} metric (Formula~\ref{form:eqop}), gender bias is the smallest, comparatively to the age and race biases.  

\section{Methodology} 
\label{sec:methodology}
This section contains detailed description of data, data preprocessing and deep learning models that were used in experiments. Data preprocessing comprises face extraction, decreasing sample size and data augmentation. There are five different architectures from two domains would be listed in total.  

\subsection{Data description}
\label{sec:sase-fe}

The database from University of Tartu has been taken as a target database for research. It contains video on which people are displaying emotions. Despite other databases for FER, frame resolution is quite high: 1280 x 960. There are 50 persons and 12 videos recorded per each, therefore in total 600 videos. Each video has high frame rate, which is equal to 100 frame per second (FPS). As was mentioned before, there are six types of emotions: happiness, surprise, sadness, disgust, anger and contempt. Originally, this database have a bit different purpose, therefore, it has additional feature - videos are divided into the genuine and fake expressions. In other words, from 12 videos, on 6 a subject is expressing true emotion, on other 6 - fake. 

\begin{figure}[hbpt]
    \begin{tabular}{c c}
        \includegraphics[width=0.23\textwidth]{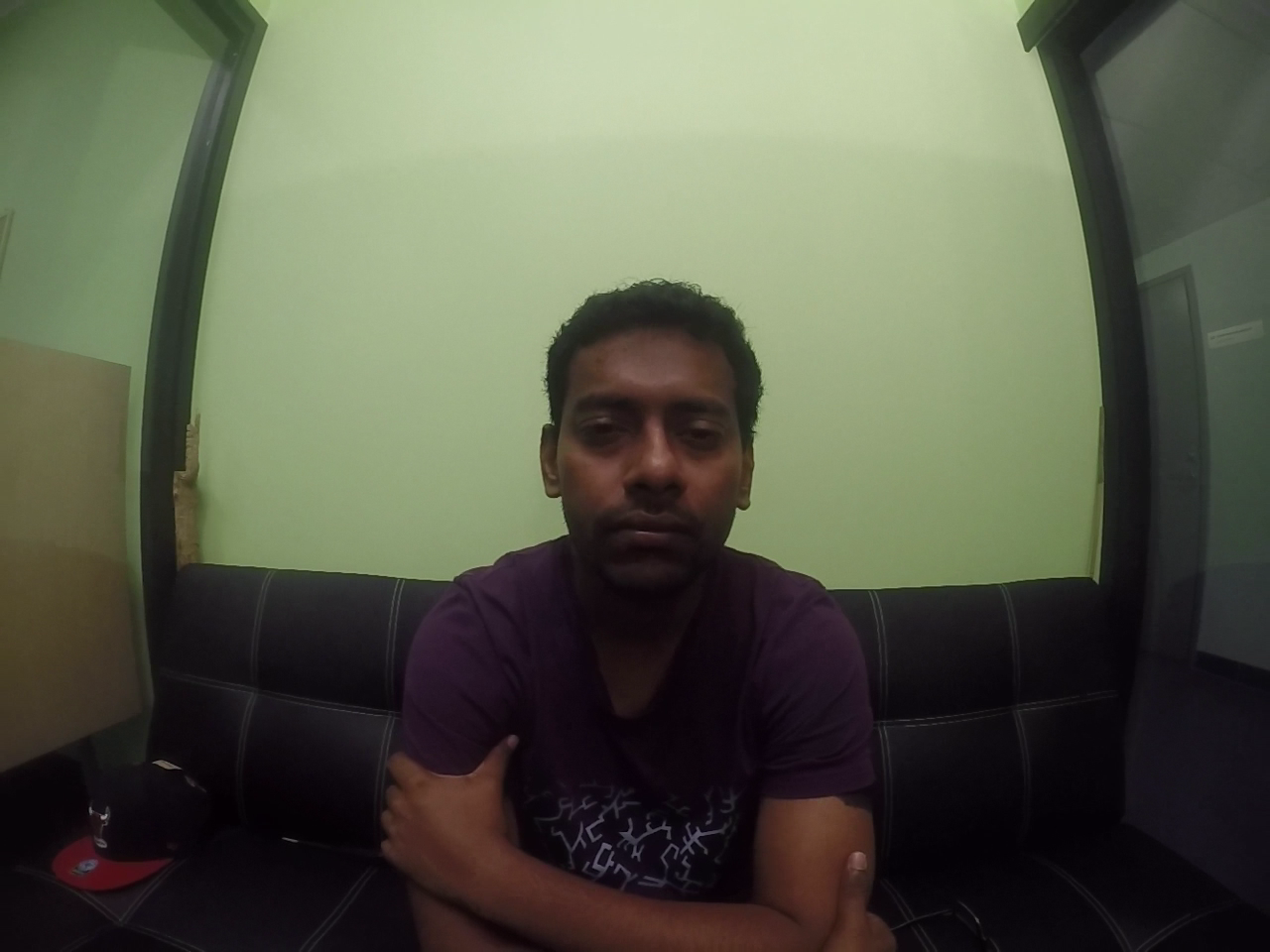} & 
        \includegraphics[width=0.23\textwidth]{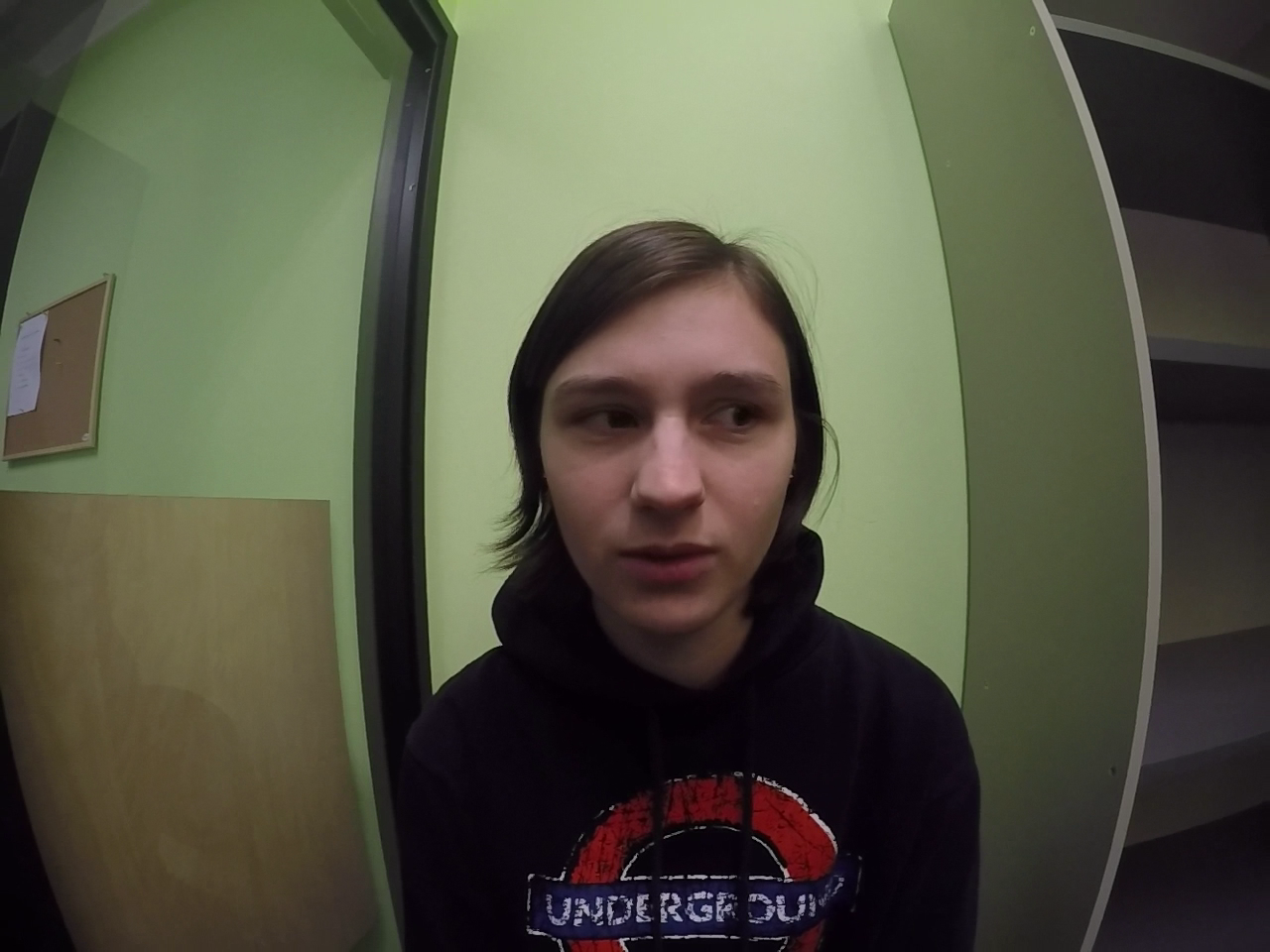} \\
        \includegraphics[width=0.23\textwidth]{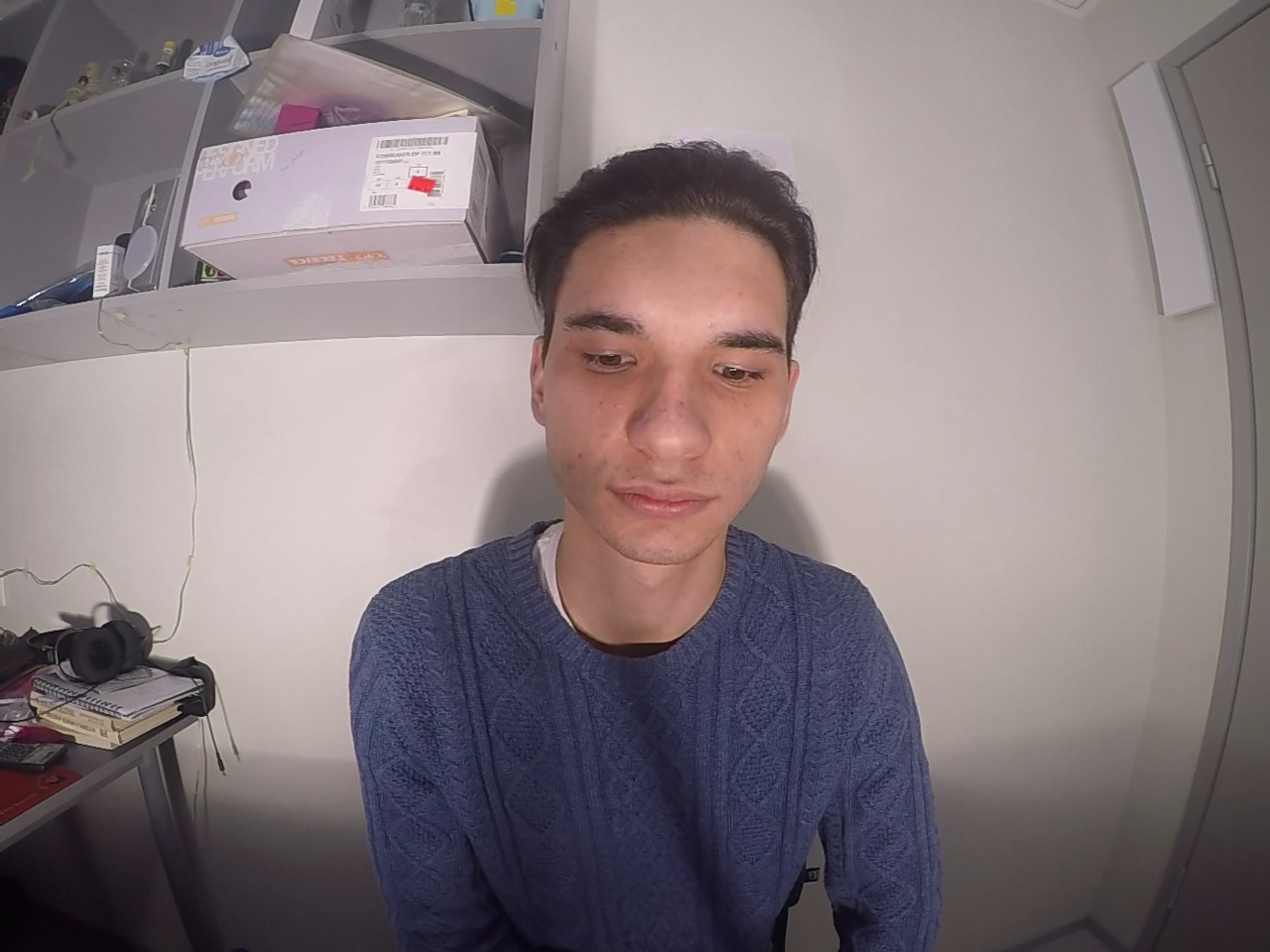} & 
        \includegraphics[width=0.23\textwidth]{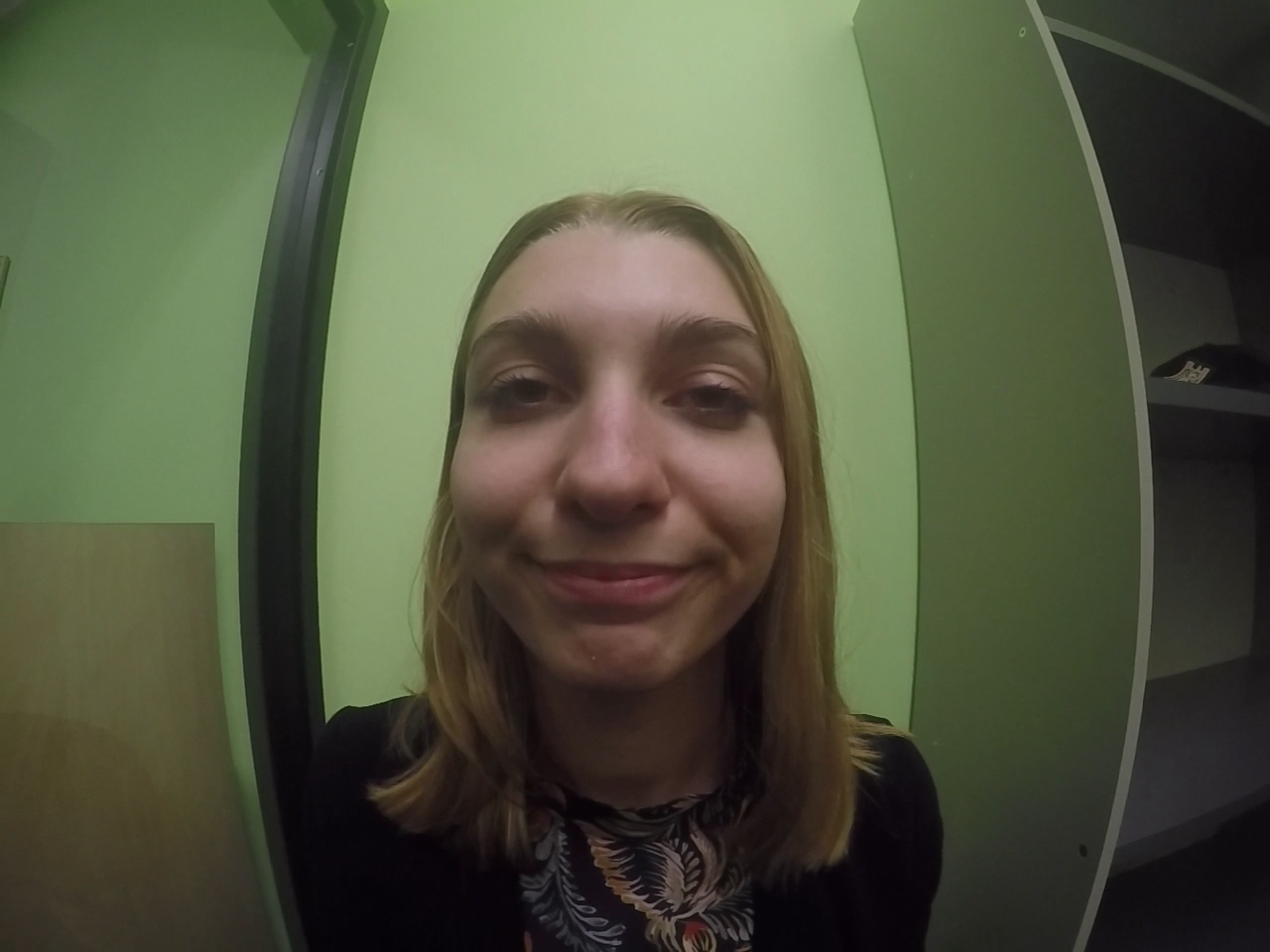}
    \end{tabular}
        \centering

        \caption{SASE-FE database examples}
        \label{fig:sase-fe}
    \end{figure}

Displaying of arbitrary emotion is always started from neutral emotion, following a command and expression which was asked. Unfortunately, there are no timestamp labels for the beginning and end of emotion, what is a considerable issue and approach to solving it will be explained further. Some samples from the dataset are shown on Figure~\ref{fig:sase-fe}.
    
\subsection{Data preprocessing} 

\subsubsection{Face extraction}

Usually, to perform face extraction, researchers use default method to extract faces by adopting Histogram of Oriented Gradients (HOG) descriptors \cite{dalal2005histograms}. DLIB \cite{dlib} implementation has been used for this task. However, it does not work well for particular dataset. Because of rich representation of skin color in dataset, the given face extractor could not extract faces for all subjects. More precisely, a race bias has been faced, since even with histogram equalization, it was not possible to extract faces for people with dark skin color. Hence, it has been decided to use another, more advance, approach - Multitask Cascade Convolutional Networks (MTCNN) \cite{xiang2017joint}. Using this model, four types of dataset were generated, each of them has different margin (0, 25, 65, 40). The size of the cropped frame is $256\times256$.
    
    \begin{figure}[hbpt]
    \begin{tabular}{c c c c}
        \includegraphics[width=0.15\textwidth]{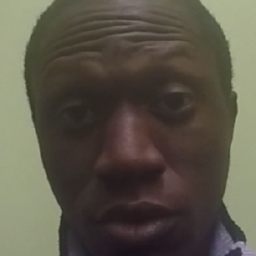} & 
        \includegraphics[width=0.15\textwidth]{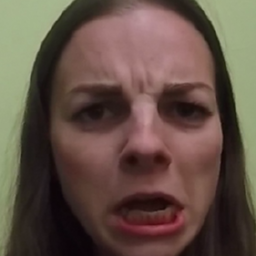}\\
        \includegraphics[width=0.15\textwidth]{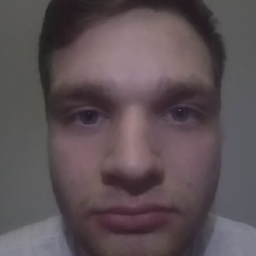} & 
        \includegraphics[width=0.15\textwidth]{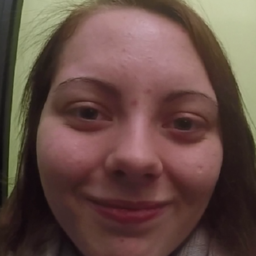}
    \end{tabular}
        \centering

        \caption{Examples of cropped images}
        \label{fig:crop-samples}
    \end{figure}

\subsubsection{Decreasing size}
\label{subsubsec:decreasing_size}

    Since database presented as video with high FPS rate, it was chosen to decrease number of frames to fit GPU capacity on High Performance Cluster (HPC). To decrease number of frames without loss of the information, K-Means \cite{lloyd1982least, macqueen1967some} algorithm was used. Simple yet effective, it allows to determine $K$ most distinguishable frames for each video. For the test purposes, it was generated $K={10, 20, 50}$. For $K=10$ the entire video was processed. For $K=20$, first 100 frames (1 sec) are excluded. For $K=50$, 20 frames from the first half of the video and 30 from the second.
    \begin{figure}[hbpt]
    \setlength{\tabcolsep}{0.1em}
    \begin{tabular}{c c c c c c c c c c}
        \includegraphics[width=0.095\textwidth]{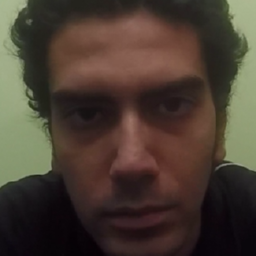} & 
        \includegraphics[width=0.095\textwidth]{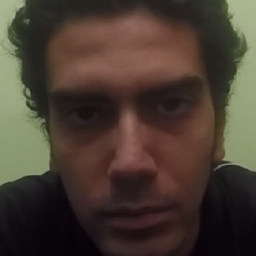} & 
        \includegraphics[width=0.095\textwidth]{ 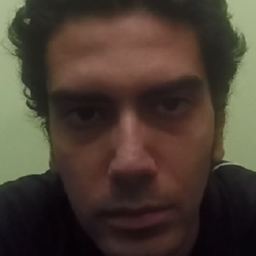} & 
        \includegraphics[width=0.095\textwidth]{ 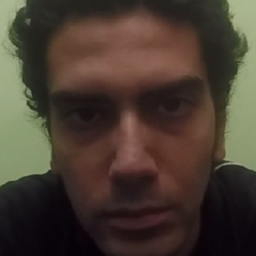} & 
        \includegraphics[width=0.095\textwidth]{ 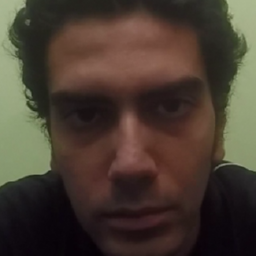} \\ 
        \includegraphics[width=0.095\textwidth]{ 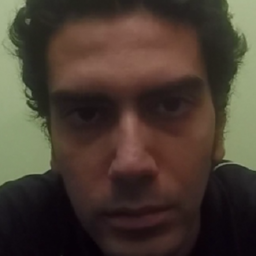} & 
        \includegraphics[width=0.095\textwidth]{ 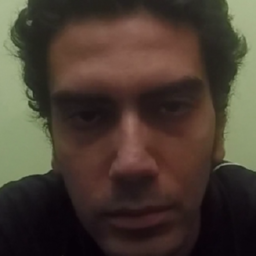} & 
        \includegraphics[width=0.095\textwidth]{ 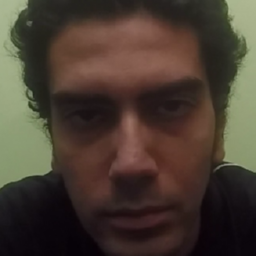} & 
        \includegraphics[width=0.095\textwidth]{ 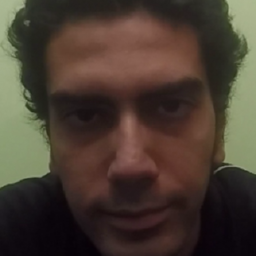} & 
        \includegraphics[width=0.095\textwidth]{ 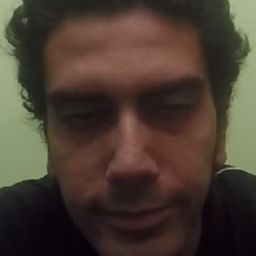} \\
        \includegraphics[width=0.095\textwidth]{ 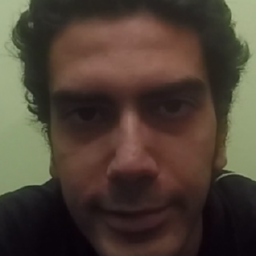} & 
        \includegraphics[width=0.095\textwidth]{ 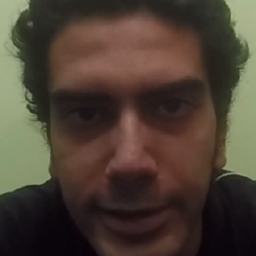} & 
        \includegraphics[width=0.095\textwidth]{ 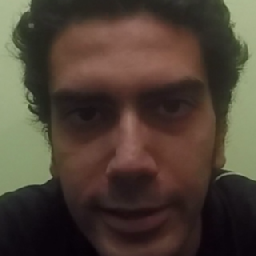} & 
        \includegraphics[width=0.095\textwidth]{ 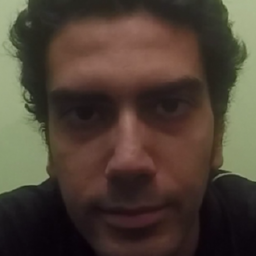} & 
        \includegraphics[width=0.095\textwidth]{ 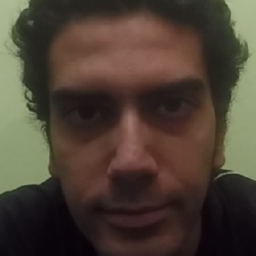} \\
        \includegraphics[width=0.095\textwidth]{ 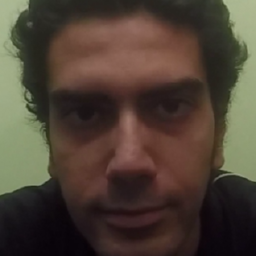} & 
        \includegraphics[width=0.095\textwidth]{ 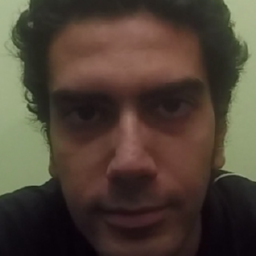} & 
        \includegraphics[width=0.095\textwidth]{ 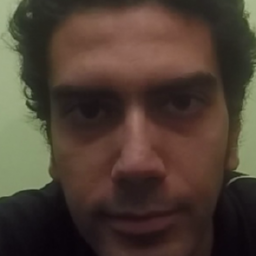} & 
        \includegraphics[width=0.095\textwidth]{ 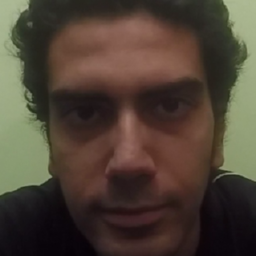} & 
        \includegraphics[width=0.095\textwidth]{ 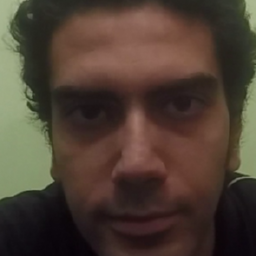} \\
    \end{tabular}
        \centering

        \caption{Example of emotion display through sequence of frames}
        \label{fig:transition}
    \end{figure}

\subsubsection{Augmentation}
The number of videos in database is relatively small. Since deep learning approaches operates better on large scale of data, data augmentation has been applied during training.

In total, there are three augmentation techniques applied:
\begin{enumerate}
    \item Horizontal flip
    \item Random rotation 
    \item Brightness augmentation
\end{enumerate}

\begin{figure}[hbpt]
    \begin{tabular}{c c c c}
        \includegraphics[width=0.15\textwidth]{ 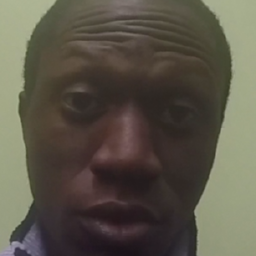} & 
        \includegraphics[width=0.15\textwidth]{ 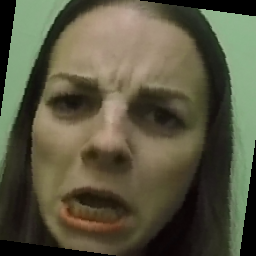} \\
        \includegraphics[width=0.15\textwidth]{ 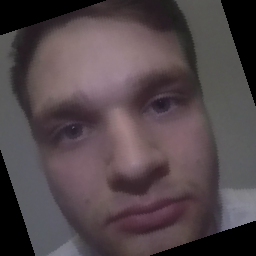} & 
        \includegraphics[width=0.15\textwidth]{ 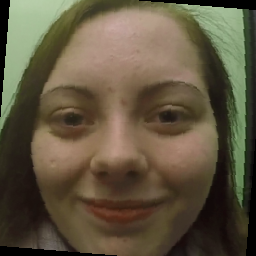}
    \end{tabular}
        \centering

        \caption{Examples of augmented data}
        \label{fig:augmented-samples}
    \end{figure}

Each augmentation is applied independently within $p=0.5$ probability on each sample. Figure~\ref{fig:augmented-samples} shows example of augmented train data.

\subsection{Deep learning models for FER}
\label{sec:using_ref}

\subsubsection{Common layers}

\paragraph{Convolution layer}
Convolution layer is a central building block for deep neural network for image data. Basically, this layer implement convolution operation, i.e. convolves the input image with given kernel. Depends whether a layer works on 2D or 3D data, convolution operation is performed along 2 or 3 dimension respectively. In case of 2D, a kernel moves along spatial dimension, meanwhile in 3D case along temporal dimension too. Implementation details and parameters can be found in \cite{Pytorch_doc}.

\paragraph{Batch normalization}
Essential presented in \cite{ioffe2015batch}, this layer helps models to reduce impact of random weight initialization, allowing to converge faster and to avoid unstable gradients. In fact, batch normalization layer only standardizes the output of the layer, by shifting and scaling with mean and standard deviation respectively. Implementation details of PyTorch BatchNorm2d and BatchNorm3d can be found in \cite{Pytorch_doc}

\paragraph{Rectified linear unit layer}
The simplest layer withing single function: to shrink all negative values, replacing them with zeros. In other words, applying function $x = max(0, x)$.

\paragraph{Pooling layer}
Pooling layer is somewhat similar to the convolution layer, however, does not contain any weights, instead just applies specific operation to the perception map. As a results, Max Pooling Layer and Average Pooling Layer are named after the function they apply. In general, pooling layer produces a single value by applying a function (maximum or average) to the perception field, sliding over input tensor dimensions. Detail information and exact implementation can be found in \cite{Pytorch_doc}

\begin{figure}[htbp]
    \centering
    \includegraphics[width=0.45\textwidth]{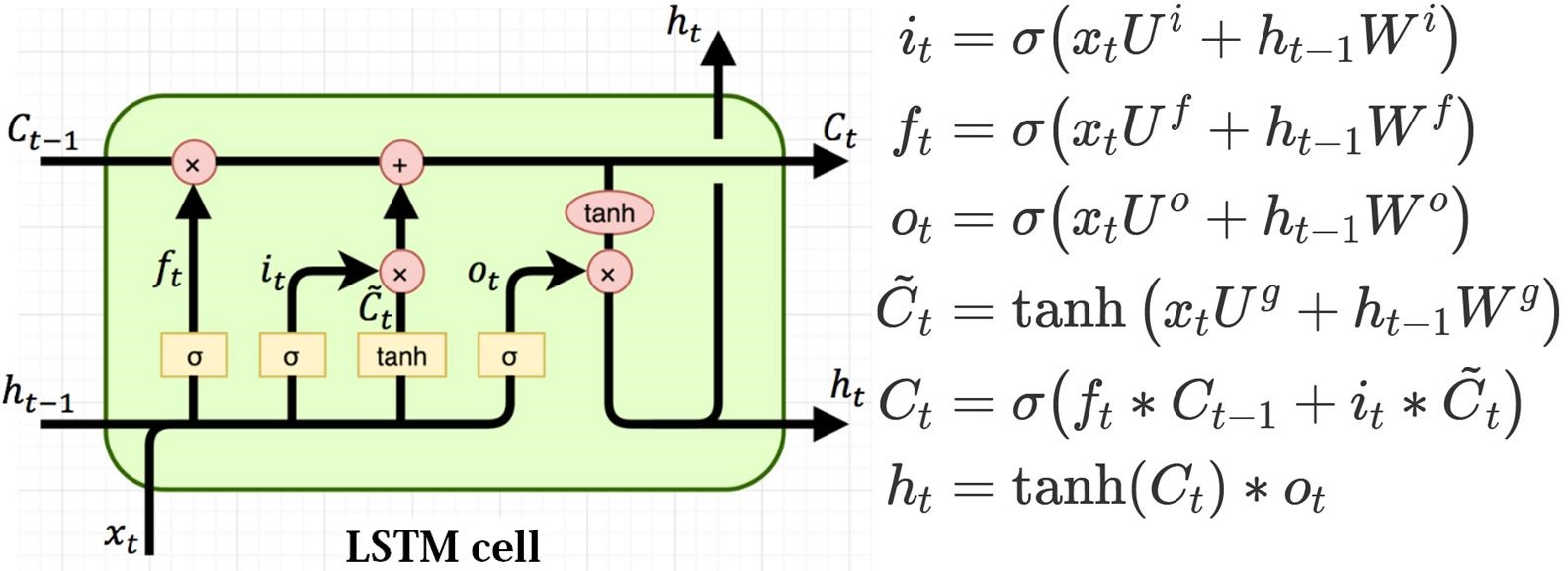}
    \caption{LSTM module \cite{varsamopoulos2018designing}}
    \label{fig:lstm}
\end{figure}

\paragraph{Long-short term memory (LSTM)}
Introduced in 90s by \cite{hochreiter1997long}, after huge success in 2013 \cite{graves2013speech}, LSTM layer became a core module for any sequence related deep learning architecture. In this paper, LSTM layer is used together with CNN feature extractors, since latter are operated only on single images, meanwhile the target input is video. LSTM itself has sophisticated structure, with 4 internal vectors named input gate($i$), output gate($o$), forget gate($f$) and cell ($C$). Along with these vectors, on each processing step LSTM contains hidden state ($h$) and cell input activation ($\Tilde{C}$) vectors. Visual explanation of how LSTM module operates is shown on Figure~\ref{fig:lstm}, where $t$ is time step, $U$ and $W$ are parameters matrix, which have to be learn.
Implementation details could be found in \cite{Pytorch_doc}.

\paragraph{Fully Connected layer (FC)}
Often the final layer, the only purpose of which is to apply affine transform $y = W\times x + b$, where $x$ is input vector, $y$ is output vector, $W$ weights and $b$ bias. Exact implementation which has been used could be found in \cite{Pytorch_doc}.

\subsubsection{VGG-LSTM}
\begin{figure}[htbp]
    \centering
    \includegraphics[width=0.49\textwidth]{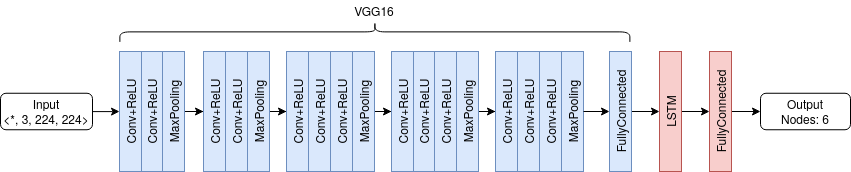}
    \caption{Architecture of VGGLSTM. Blue: VGG16 backbone with pretrained weights. Red: layers trained from the scratch.}
    \label{fig:vgg}
\end{figure}

One of the most famous model for face recognition which has been widely used and has become a background for further research is VGG16 architecture \cite{parkhi2015deep}, named after Visual Geometry Group, who has conducted research. In this work, VGG16 model serve as a feature extractor which preceded LSTM block. Together with a fully connected layer at the end, they create a model for classification sequential data. The backbone model consists of series of blocks, which are two or three convlutional layers with ReLU and the MaxPooling layer on the end. The model ends with fully connected layer. Exactly this layer serves as feature vector, which further goes to LSTM module, next down to the fully connected layer. The entire structure is on Figure~\ref{fig:vgg}. As were mentioned before, VGG architecture is wide used, therefore there are many good pretrained weights available for the research. It has been chosen to use pretrained weights from ImageNet dataset \cite{deng2009imagenet} and VGGFace dataset \cite{parkhi2015deep}. To simplify further presentation of results, VGG16 architecture with pretrained ImageNet weights is denoted as \textit{VGG16}, meanwhile VGG16 architecture with pretrained VGGFace weights is denoted as \textit{VGGFACE}.

\subsubsection{ResNet50-LSTM}
\begin{figure}[htbp]
    \centering
    \includegraphics[width=0.45\textwidth]{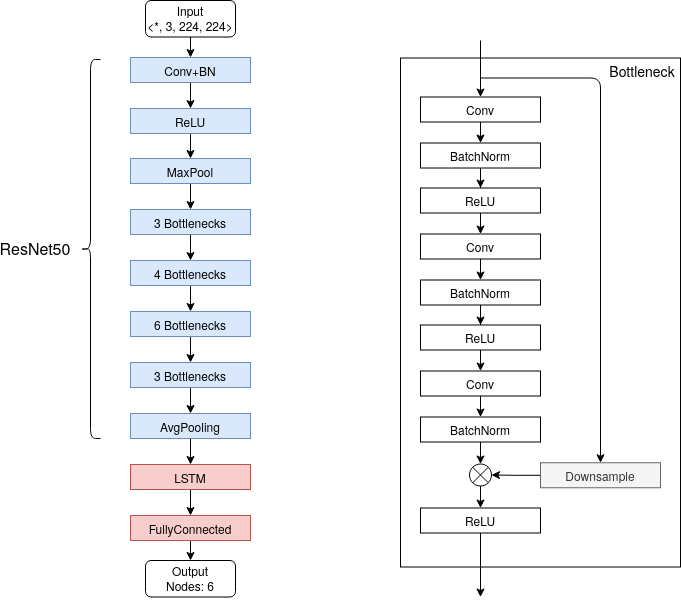}
    \caption{Architecture of ResNet50-LSTM. Blue: ResNet50 backbone with pretrained weights. Red: layers trained from the scratch.}
    \label{fig:resnet50}
\end{figure}

Another great example of a model, which is a result of consecutive research in deep learning is ResNet50 \cite{he2016deep}. Following the same principle in construction, the model is presented as a CNN feature extractor. ResNet50 backbone begins with a sequence of convolutional layer, batch normalization, ReLU and MaxPooling. Next, four layers with 3, 4, 6 and 3 Bottleneck blocks respectively. Model ends with AvgPooling layer to reduce output feature map size. At this point there are no more backbone layers, and data stream goes down to LSTM module, following by fully connected layer. The entire architecture is visualised on Firuge~\ref{fig:resnet50}. Since the model is too deep by itself, ImageNet pretrained weights are used. In this paper, when we are referring to \textit{ResNet50}, meaning entire architecture, which consists of a backbone and LSTM module. 

\subsubsection{SENet-LSTM}
\begin{figure} [htbp] 
\begin{center}
\includegraphics[width=0.45\textwidth]{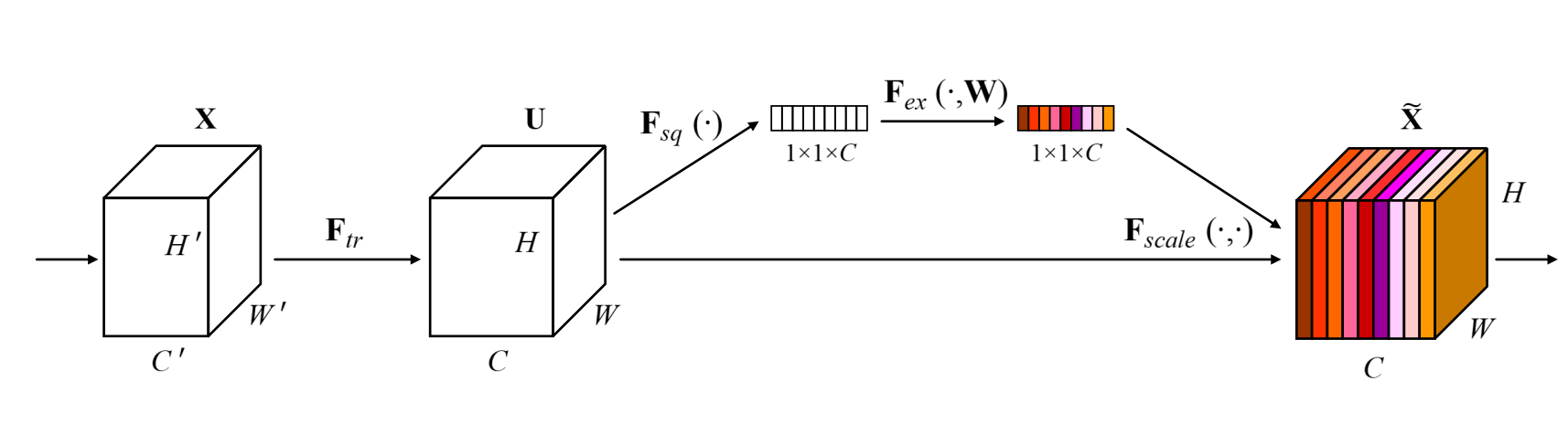}
\caption{Squeeze-and-Excitation block \cite{hu2018squeeze}}
\label{fig:squeeze_exc}
\end{center}
\end{figure}

In \cite{hu2018squeeze} authors discovered, that integrating mechanism for learning relationship between channels can considerably improve results. The goal of such block is to explicitly train a network the cross-correlation between spatial channels. The main advantages of this approach are lightweight, simplicity to implement and application to all common layers. Basically, for any given transformation $F_{tr}: X \xrightarrow{} U$, first, squeeze the feature map into the descriptor vector and then multiply it to the feature block channel wise.
See Figure~\ref{fig:squeeze_exc}.

In general, architecture of SENet-LSTM consists of two parts: a backbone and LSTM module. The backbone is SENet model, which copies the architecture of ResNet50 (blue blocks on Figure~\ref{fig:resnet50}), however, each bottleneck block enhanced with SE feature. In other words, Bottleneck layer plays role of function $F_{tr}$, according to the Figure~\ref{fig:squeeze_exc}. The LSTM module is a one layer LSTM layer withing one fully connected follow up layer. SENet backbone has 8096 output features. LSTM block has 256 hidden neurons. Since the number of training data is quite small, it has been chosen to use pretrained weights for a backbone. Ready to use weights, trained of VGGFace2 dataset\cite{cao2018vggface2} for SENet was taken from \cite{senetgithub}. For purpose of convenience, further this architecture is referred as \textit{SENetLSTM}.

\subsubsection{3D-CNN}
\begin{figure}[htbp]
    \centering
    \includegraphics[width=0.45\textwidth]{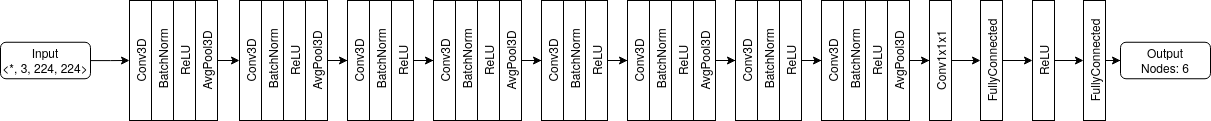}
    \caption{Architecture of 3D-CNN model}
    \label{fig:3dcnn}
\end{figure}

Among others approaches which utilize 3D CNN \cite{zhao2018learning, reddy2019spontaneous} for FER, in \cite{haddad20203d} authors propose an architecture within optimized hyperparameters for CK+ and OULU-CASIA dataset. The network takes input 10 consecutive RGB frames $112\times112$ resolution frames. However, in purposes of this research work, to be consistent with other network, the input resolution has been changed to $224\times224$, and the sequence length varies. Hence, the model architecture has been preserved up to fully connected layers, since their size is directly depended on the input. The final model is presented on Figure~\ref{fig:3dcnn}. This model does not have any pretrained weights and trained from the scratch. In the further sections this model is referred as \textit{3DCNN}.

\subsubsection{ResNet3D}
\begin{figure}[htbp]
    \centering
    \includegraphics[width=0.45\textwidth]{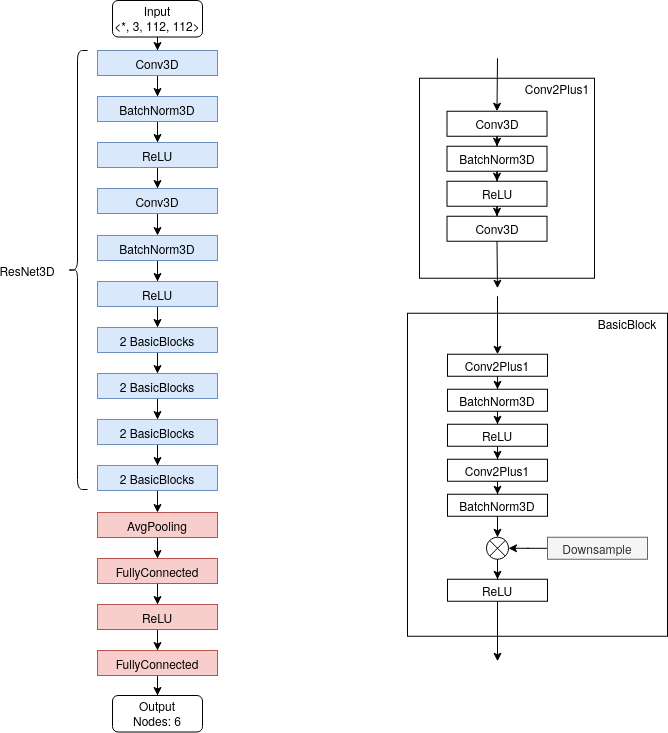}
    \caption{Structure of ResNet3D-LSTM model}
    \label{fig:resnet3d}
\end{figure}

In 2018, a group of researchers proposed a new way to treat 3D data such as videos, presenting another two convolutional blocks: mixed convolutions and "(2+1)D" convolutions \cite{tran2018closer}. Utilizing these block, authors constructed ResNet3D, an analogue of already famous ResNet2D. Defining the clone of shorted version of ResNet - ResNet18 - ResNet3D has got three different implementations: with 3D convolutions, with mixed convolutions and with "2+1"D convolutions. Since authors claim that the best performance has been demonstrated by "2+1"D convolutions, in this paper exactly this type of model is used. The idea behind "2+1"D convolutions is to think about 3D convolutions as 2D convolutions in spatial space, which are followed by 1D convolutions in the temporal space. The benefits from these approach are following: first, due to the additional ReLU after 2D convolutions, this block has twice more number of nonlinearities, increasing capacity of the model, second, authors claim that during the training, a tensor wise error rate is smaller, compared to the 3D convolution counterparts. The overall architecture of ResNet3D is similar to ResNet50, but with different blocks. The final model structure is presented on Figure~\ref{fig:resnet3d}. 

\subsection{Task definition}
The goal is to train a classifier to classify an emotion out of six classes: surprise, contempt, disgust, anger, sad and happy. Let us define a neural network as probabilistic classifier $P$, with parameters $\Theta$. In this paper, a classification is performed in the "soft" form, i.e.
\begin{equation}
    \hat{y} = Softmax_{y}(P(Y=y|X, \Theta))
\end{equation}
where $X$ is an input, $Y$ is an output, $y \in {0,1,2,3,4,5}$ according to the six type of emotions. Softmax, in turn, is defined as follows
\begin{equation}
    Softmax(x_{i}) = \frac{e^{x_i}}{\sum^{5}_{j=0} e^x_j}
\end{equation}

\section{Experiments}
\label{sec:experimetns}
The following section has an overview of all details regarding experiments. Hence, below will be described data division and hyperparameters. In addition, this section contains description of necessary metrics, based on which conclusion would be derived.

\subsection{Input size}

The raw instance of database is a video file. Due to high resolution of videos and limited computational capacity, as was mentioned in~\ref{subsubsec:decreasing_size}, each video has been squeezed for particular number of frames. However, the observations are that: within $K=20$ models generalize data the best. Using $K=10$, there were not any adequate results and with $K=50$ there were not observed any increase of in performance. Moreover, due to the limited computational resources, bigger length of the input leads to decreasing of batch size, consequently increasing training time or even not allowing to fit a model. The input resolution is $(224\times 224)$ for all models, except \textit{ResNet3D}. Due to the enormous number of weights, available computational capacity is not enough to fit the model. Hence, the input resolution for \textit{ResNet3D} is $(112\times 112)$. Putting all together, the input sample for \textit{ResNet3D} is $(10, 3, 112, 112)$ and for all other models is $(10, 3, 224, 224)$.

\subsection{Data division}

SASE-FE dataset contains 18 female and 32 male subjects. The instances for the test group have been selected randomly and once. Overall we had selected 5 males and 5 females. Hence, all videos corresponded to these persons are not visible anyhow during the training. Since each person has 12 videos, therefore test set contains 120 videos, 20 videos per emotion. 

Often, during training process a model tends to underfit or overfit. To track these kinds of behavior, making training process more efficient and to pick the best model, it has been decided to organize a validation test with fixed size as 15\% of train data. A validation set is generated randomly for each train run. 

To sum up, each training process contains train and validation sets with 32 and 8 subjects correspondingly. Multiplying by 12, there are 384 train videos and 96 validation videos. Test set with total 120 videos is not involved during training process. The desired metrics which are related to the goals of this paper are calculated exactly from inference on the test set.

\subsection{Training details}

All training experiments were performed on University of Tartu HPC. The target GPU is NVIDIA Tesla-V100 with 32 GB of VRAM. All code is written in Python with usage of machine learning python package - PyTorch. The hyperparameter search during the entire research was consisted of two phases for each model. First phase of hyperparameter search was performed to find out the optimal model, hence target hyperparameters were such things as number of fully connected layers, number of LSTM layers, number of nodes in these layers. After finding an optimal structure, in the second phase the optimal set of training hyperparameters for each model have been discovered. In the scope of experiments, such variables have been varied: learning rate, batch size and weight decay. The final hyperparameters are presented in Table~\ref{tab:hyperparameters}.

\begin{table}[H]
    \centering
    \begin{tabular}{|c|c|c|c|}
    \hline
        Model &    Learning rate  &  Batch size & Weight decay  \\
\hline
        \textit{3DCNN} & 5e-3 & 12 & 1e-6 \\
        \textit{ResNet3D} & 5e-3 & 12 & 5e-3 \\
        \textit{ResNet50} & 5e-3 & 12 & 1e-3 \\
        \textit{SENetLSTM} & 1e-2 & 12 & 5e-4 \\
        \textit{VGG16} & 1e-2 & 12 & 5e-3 \\
        \textit{VGGFACE} & 5e-3 & 12 & 8e-4 \\
\hline
    \end{tabular}
    \caption{Hyperparameters values for each model}
    \label{tab:hyperparameters}
\end{table}

In order to perform gender bias analysis, test set has been divided into two subsets: pure male set and pure female set, such that each has 5 subjects. In this scenario, pure means to include subjects of only one gender. Hence, from here and further, there are three sets for testing, named as follows: \textbf{Test set}, \textbf{Male set} and \textbf{Female set}. Moreover, it was decided to look how performance and metrics differ depends on training data. For this purpose, all train data also was divided into pure male train set and pure female train set. Therefore, to follow results easier, models which are trained on entire train set are called \textbf{Regular}, trained on pure male set and pure female set are called \textbf{Male} and \textbf{Female} respectively. Training of all models have been performed using Stochastic Gradient Decent (SGD) optimization. To sum up, it has been trained 18 different neural networks, 6 different models by 3 different train sets. 

\subsection{Metrics}

According to the previously mentioned definitions of fairness, the main metrics, based on which any conclusion could be made are accuracy, true positive rate (TPR) and false positive rate (FPR). Although these metrics are trivial, formal definitions are listed below: 

\begin{equation}
    ACC = \frac{TP+NT}{TP+TN+FP+FN}  
\end{equation}
\begin{equation}
    TPR = \frac{TP}{TP+FN}
\end{equation}
\begin{equation}
    FPR= \frac{FP}{FP+TN}
\end{equation}
where $TP$ - true positive, $FP$ - false positive, $TN$ - true negative, $FN$ - false negative, $ACC$ - accuracy, $TPR$ and $FPR$ - true positive and false positive rate respectively.

The first definition of fairness (Equation~\ref{form:eqod}) implies the equality of TPR and FPR simultaneously. The second definition (Equation~\ref{form:eqop}), extending to the multi classification case, requires equal TPR for the groups with different unprotected attribute value. And the third, and the last, definition of fairness (Equation~\ref{form:dp}) means the equal accuracy for separated groups. In other words, the difference in these metrics shows how fair model is, having linear dependency - the higher discrepancy, the more biased model is.

Although target database comprises six different emotions, usually it is extremely difficult to distinguish between contempt, disgust and anger. In order to relax constraints and make classification a bit easier, without losing general idea, it has been decided to fuse these three emotions into a single emotion and name it "Upset". 

\section{Results}
\label{sec:results}
This section consists of 2 parts. First part contains confusion matrices and shallow analysis of each. Second part is served for more detailed and precise analysis from the perspective of fairness, emotions and test sets.

\subsection{Confusion matrices}
\begin{figure}[!h]
\begin{tabular}{c c c}
\includegraphics[width=0.15\textwidth]{ 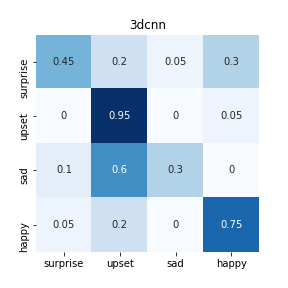}
&
\includegraphics[width=0.15\textwidth]{ 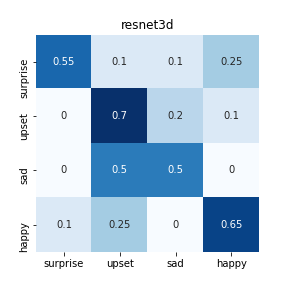}
&
\includegraphics[width=0.15\textwidth]{ 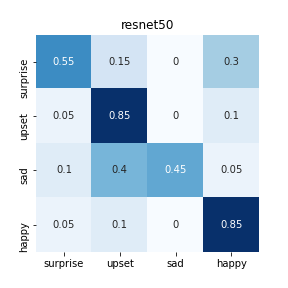}\\
 \includegraphics[width=0.15\textwidth]{ 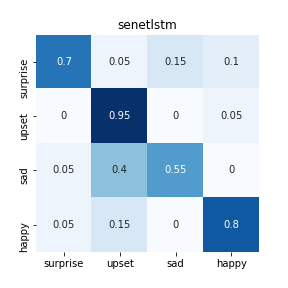}
 &
\includegraphics[width=0.15\textwidth]{ 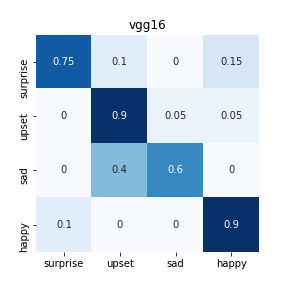}
&
\includegraphics[width=0.15\textwidth]{ 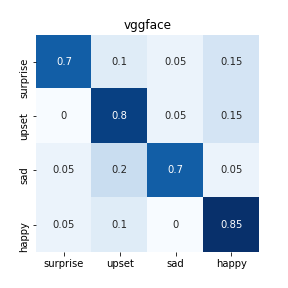}\\
\end{tabular}
\caption{Confusion matrix per method for Regular models on Test set}
\label{fig:rt_conf_matr}
\end{figure}

Regular models have shown decent performance on Test set. The major part of emotions are classified correctly, however, Upset and Sad emotions are frequently misclassified, which is an expected outcome, since they are close in V-A space. Confusion matrices are shown on Figure~\ref{fig:rt_conf_matr}.

\begin{figure}[!h]
\begin{tabular}{c c c}
\includegraphics[width=0.15\textwidth]{ 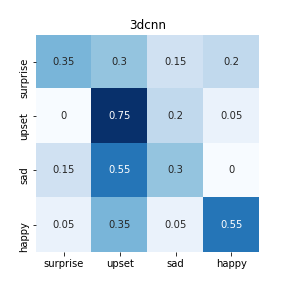}
&
\includegraphics[width=0.15\textwidth]{ 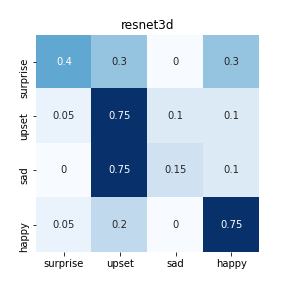}
&
\includegraphics[width=0.15\textwidth]{ 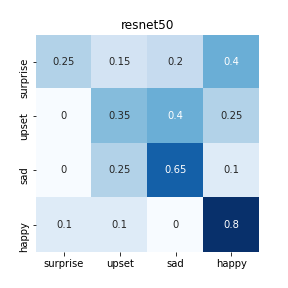}\\
\includegraphics[width=0.15\textwidth]{ 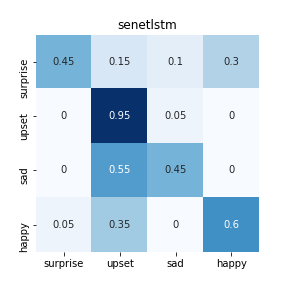}
&
\includegraphics[width=0.15\textwidth]{ 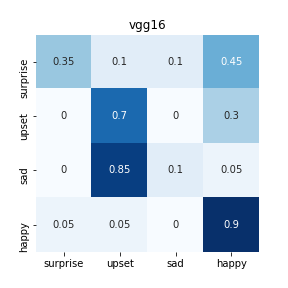}
&
\includegraphics[width=0.15\textwidth]{ 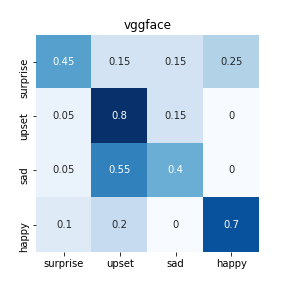}\\
\end{tabular}
\caption{Confusion matrix per method for Female models on Test set}
\label{fig:ft_conf_matr}
\end{figure}

A decent decrease in accuracy for Surprise, which indicates that female training data is lack of Surprise expression. Confusion matrices are on Figure~\ref{fig:ft_conf_matr}. 

\begin{figure}[!h]
\begin{tabular}{c c c}
\includegraphics[width=0.15\textwidth]{ 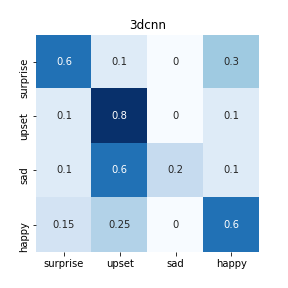}
&
\includegraphics[width=0.15\textwidth]{ 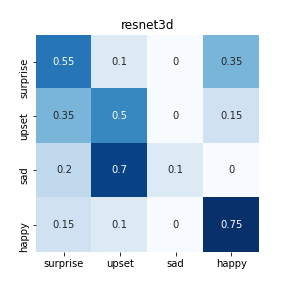}
&
\includegraphics[width=0.15\textwidth]{ 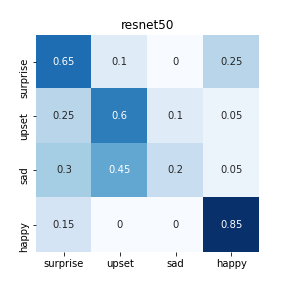}\\
\includegraphics[width=0.15\textwidth]{ 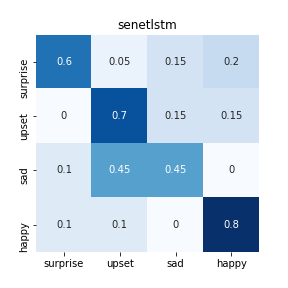}
&
\includegraphics[width=0.15\textwidth]{ 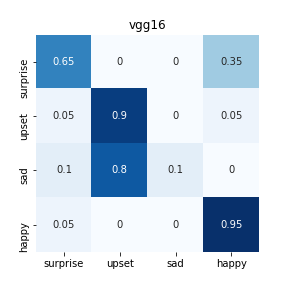}
&
\includegraphics[width=0.15\textwidth]{ 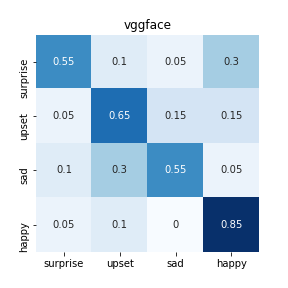}\\
\end{tabular}
\caption{Confusion matrix per method for Male models on Test set}
\label{fig:mt_conf_matr}
\end{figure}

We can observe almost the same performance comparatively to the Regular models, however, the misclasification rate of Sad emotion as Upset is much higher. This implies that male training data includes more Sad samples which are visually much closer to Upset emotion. Confusion matrices are on Figure~\ref{fig:mt_conf_matr}.

\begin{figure}[!h]
\begin{tabular}{c c c}
\includegraphics[width=0.15\textwidth]{ 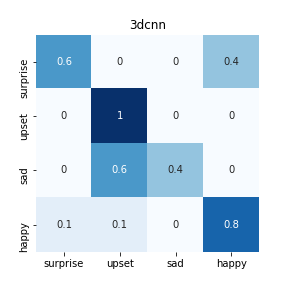}
&
\includegraphics[width=0.15\textwidth]{ 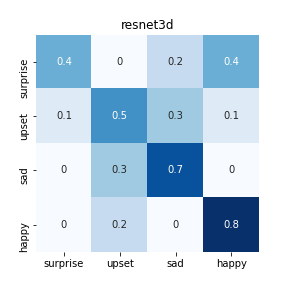}
&
\includegraphics[width=0.15\textwidth]{ 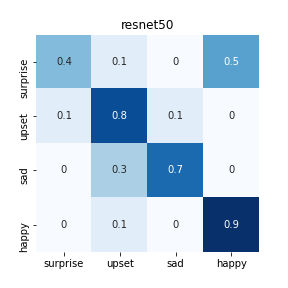}\\
 \includegraphics[width=0.15\textwidth]{ 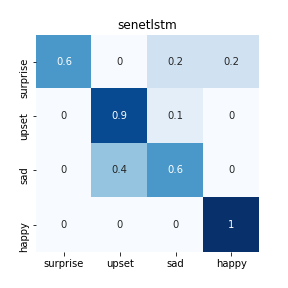}
 &
\includegraphics[width=0.15\textwidth]{ 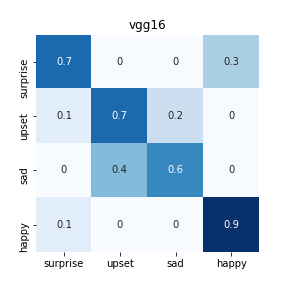}
&
\includegraphics[width=0.15\textwidth]{ 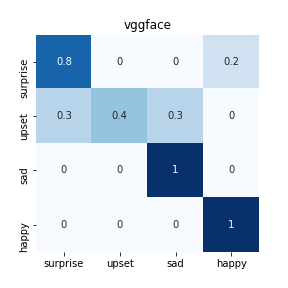}\\
\end{tabular}
\caption{Confusion matrix per method for Regular models on Female set}
\label{fig:rf_conf_matr}
\end{figure}

Regular models have shown almost excellent recognition of Happy emotion for Female set. TPR for Surprise in not high, however, we observe less misclassification between Upset and Sad. Confusion matrices are on Figure~\ref{fig:rf_conf_matr}. 

\begin{figure}[!h]
    \begin{tabular}{c c c}
    \includegraphics[width=0.15\textwidth]{ 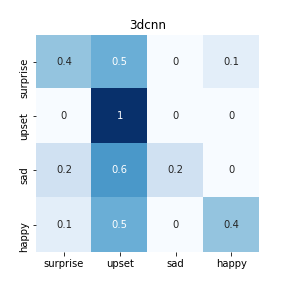}
    &
    \includegraphics[width=0.15\textwidth]{ 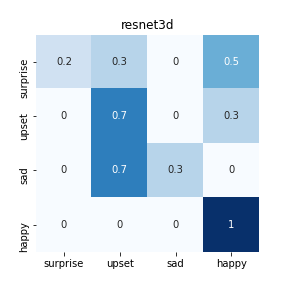}
    &
    \includegraphics[width=0.15\textwidth]{ 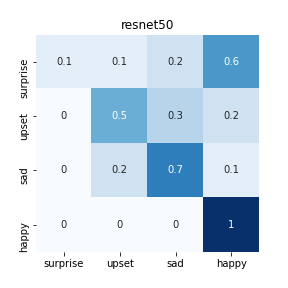}\\
    \includegraphics[width=0.15\textwidth]{ 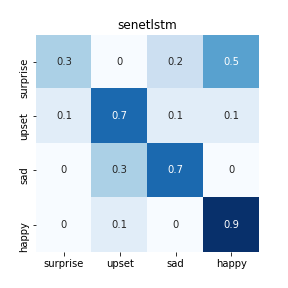}
    &
    \includegraphics[width=0.15\textwidth]{ 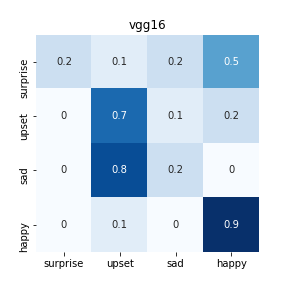}
    &
    \includegraphics[width=0.15\textwidth]{ 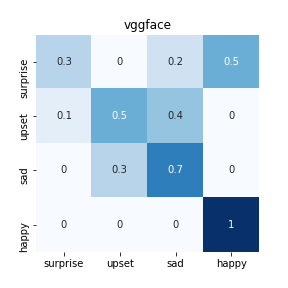}\\
    \end{tabular}
    \caption{Confusion matrix per method for Female models on Female set}
    \label{fig:ff_conf_matr}
\end{figure}

Quite unexpected results for Female models on Female set. Only TPR for Happy emotion is high enough to consider as acceptable. Recognition of Surprise is very low, which means that female training data has weak samples for Surprise. Confusion matrices are on Figure~\ref{fig:ff_conf_matr}. 

\begin{figure}[!h]
\begin{tabular}{c c c}
\includegraphics[width=0.15\textwidth]{ 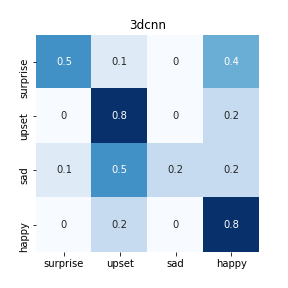}
&
\includegraphics[width=0.15\textwidth]{ 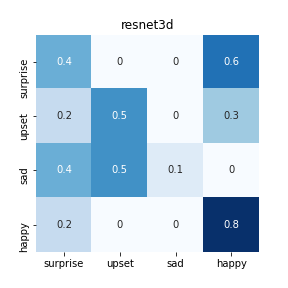}
&
\includegraphics[width=0.15\textwidth]{ 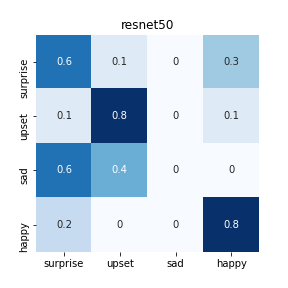}\\
 \includegraphics[width=0.15\textwidth]{ 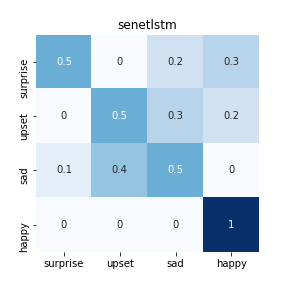}
&
\includegraphics[width=0.15\textwidth]{ 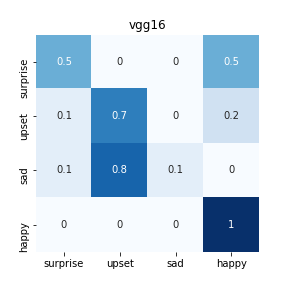}
&
\includegraphics[width=0.15\textwidth]{ 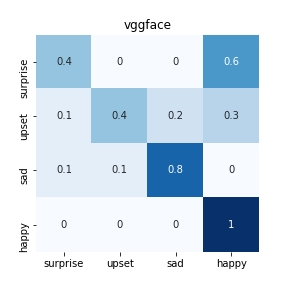}\\
\end{tabular}
\caption{Confusion matrix per method for Male models on Female set}
\label{fig:mf_conf_matr}
\end{figure}

As has been stated before, Female training data has weak Surprise samples, meanwhile Male models have shown average performance, hence, even with different gender domain, male samples of Surprise are much stronger. Confusion matrices are on Figure~\ref{fig:mf_conf_matr}.

\begin{figure}[!h]
\begin{tabular}{c c c}
\includegraphics[width=0.15\textwidth]{ 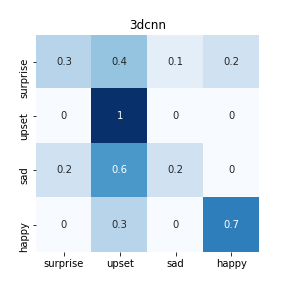}
&
\includegraphics[width=0.15\textwidth]{ 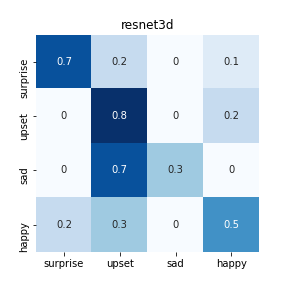}
&
\includegraphics[width=0.15\textwidth]{ 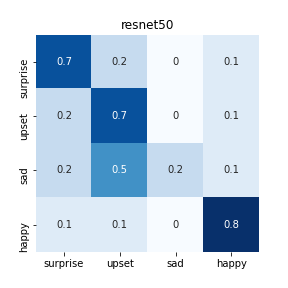}\\
 \includegraphics[width=0.15\textwidth]{ 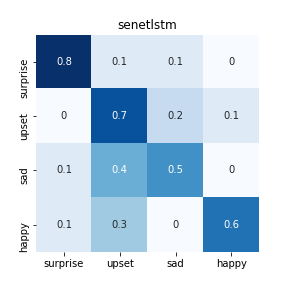}
&
\includegraphics[width=0.15\textwidth]{ 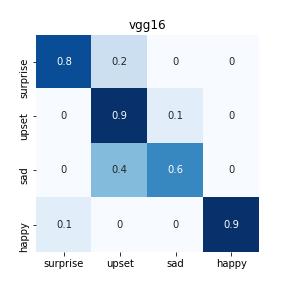}
&
\includegraphics[width=0.15\textwidth]{ 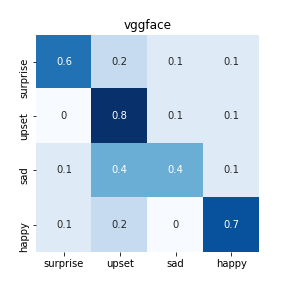}\\
\end{tabular}
\caption{Confusion matrix per method for Regular models on Male set}
\label{fig:rm_conf_matr}
\end{figure}

Comparatively to other test sets, inference of Regular models on Male set has much lower TPR for Happy emotion and much higher TPR for Surprise emotion. Confusion matrices are on Figure~\ref{fig:rm_conf_matr}.

\begin{figure}[!h]
\begin{tabular}{c c c}
\includegraphics[width=0.15\textwidth]{ 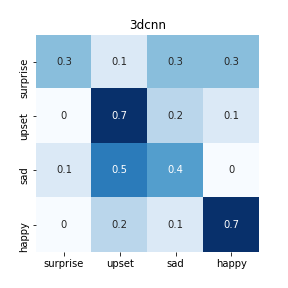}
&
\includegraphics[width=0.15\textwidth]{ 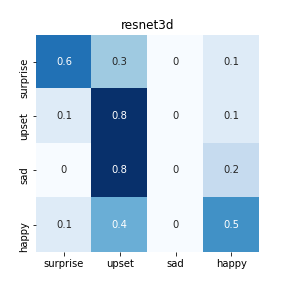}
&
\includegraphics[width=0.15\textwidth]{ 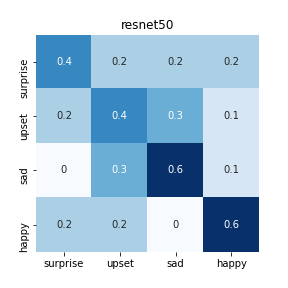}\\
\includegraphics[width=0.15\textwidth]{ 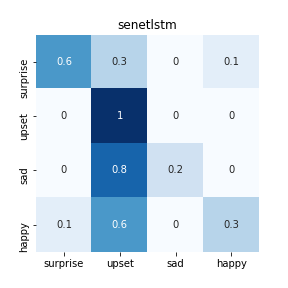}
&
\includegraphics[width=0.15\textwidth]{ 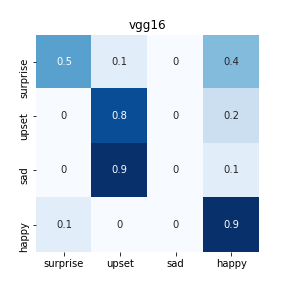}
&
\includegraphics[width=0.15\textwidth]{ 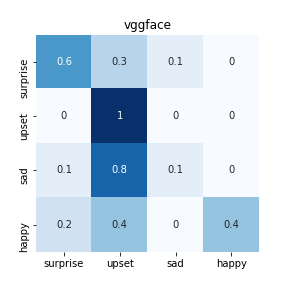}\\
\end{tabular}
\caption{Confusion matrix per method for Female models on Male set}
\label{fig:fm_conf_matr}
\end{figure}

Female models on the Male set have probably the worst metrics overall, due to the lack of data in train data and opposite gender in test data. Confusion matrices are on Figure~\ref{fig:fm_conf_matr}.

\begin{figure}[!h]
\begin{tabular}{c c c}
\includegraphics[width=0.15\textwidth]{ 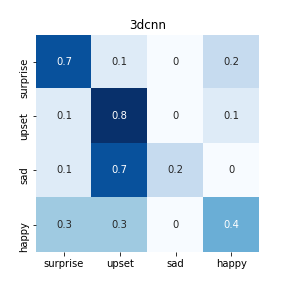}
&
\includegraphics[width=0.15\textwidth]{ 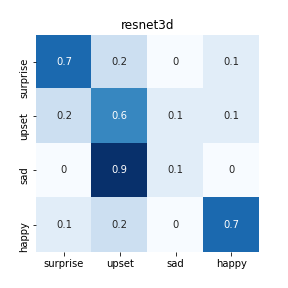}
&
\includegraphics[width=0.15\textwidth]{ 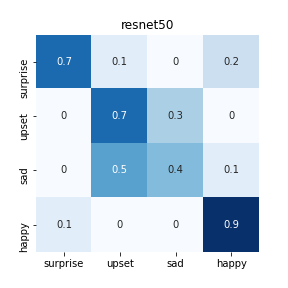}\\
 \includegraphics[width=0.15\textwidth]{ 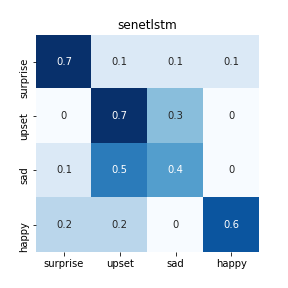}
&
\includegraphics[width=0.15\textwidth]{ 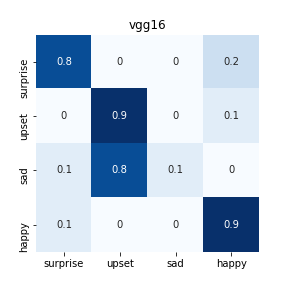}
&
\includegraphics[width=0.15\textwidth]{ 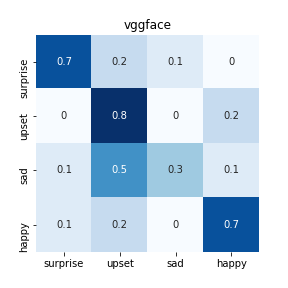}\\
\end{tabular}
\caption{Confusion matrix per method for Male models on Male set}
\label{fig:mm_conf_matr}
\end{figure}

Male models on Male set have shown great accuracy for Surprise and Happy emotions. There are high TPR for Upset, however, misclassification rate of Sad as Upset is high too. Since train and test data share the same gender, performance expectations have been higher. Confusion matrices are on Figure~\ref{fig:mm_conf_matr}.

\subsection{Analysis}

The goal of this research work is to analyse a gender bias, according to the definition of fairness. Raw results on which the following analysis is based are listed in Appendix~\ref{appendix:tables}. In the matter of convenience, several chars are presented in Appendix~\ref{appendix:charts}.

\begin{figure}[!h]
    \centering
    \begin{tabular}{c c c}
       \includegraphics[width=0.15\textwidth]{ 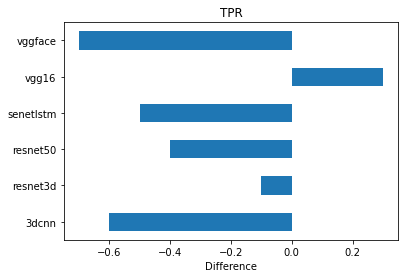} &
       \includegraphics[width=0.15\textwidth]{ 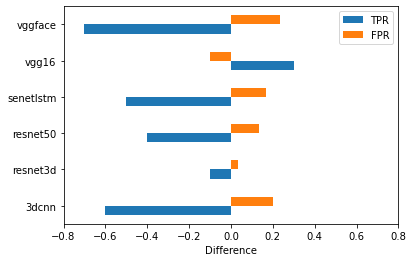} &
       \includegraphics[width=0.15\textwidth]{ 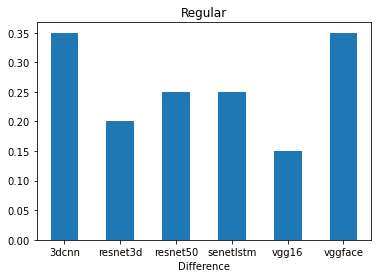}\\
       (a) & (b) & (c)\\
    \end{tabular}
    \caption{Metrics for models, which have been trained on entire train data}
    \label{fig:metrics_regular}
\end{figure}

According to equal opportunity (EQOP) and equalized odds (EQOD) (Figure~\ref{fig:metrics_regular} (a, b)), the least biased is \textit{ResNet3D}, while \textit{VGGFACE} is the most unfair. Meanwhile, \textit{3DCNN} is second in term of biasness. However, comparison in accuracy difference (demographic parity (DP)) (Figure~\ref{fig:metrics_regular} (c)) consider the most fair model \textit{VGG16}, when \textit{ResNet3D} is the second most fair one. As for the most biased models according to DP, these are \textit{3DCNN} and \textit{VGGFACE}. Hence, overall results are consider to be aligned along all three definitions of fairness.

\begin{figure}[!h]
    \centering
    \includegraphics[width=0.48\textwidth]{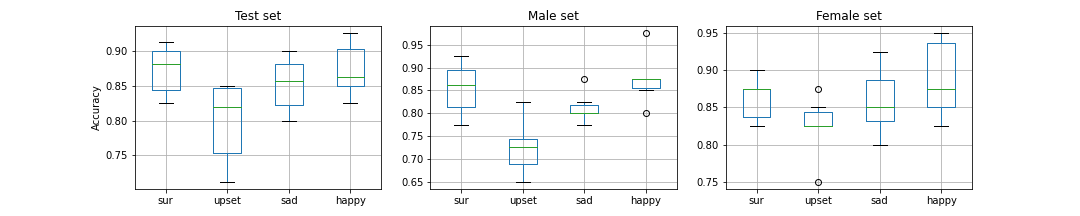}
    \caption{Aggregated accuracy for models, trained on the entire data. Comparison for three different test sets.}
    \label{fig:boxplot_regular}
\end{figure}

Analysing aggregated results (Figure~\ref{fig:boxplot_regular}), several conclusions have been derived. For the Test and Male set, Regular models show the best accuracy for classification Surprise, while on the Female set, the best accuracy is for Happy emotion. Fused Upset emotion has the lowest recognition rate for all three sets. Worth to mention, inference of Female set has much higher variance rather than on Male set. However, average accuracy for Female set is higher.

\begin{figure}[!h]
    \centering
    \begin{tabular}{c c}
       \includegraphics[width=0.23\textwidth]{ 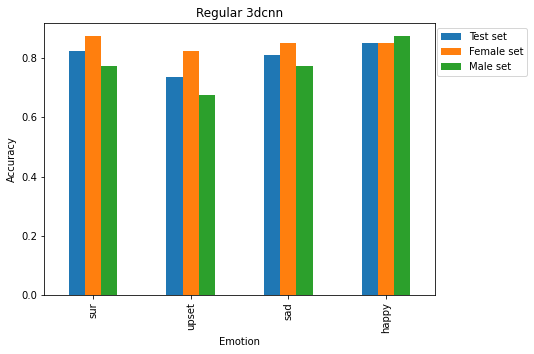}  & 
       \includegraphics[width=0.23\textwidth]{ 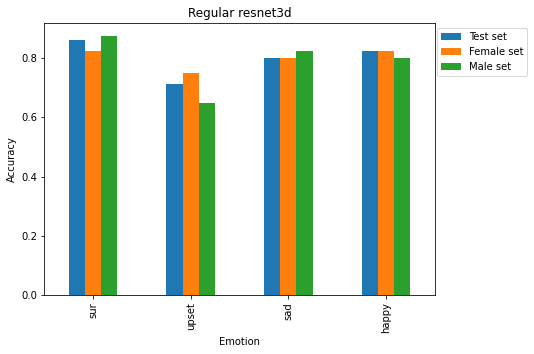}\\
       \includegraphics[width=0.23\textwidth]{ 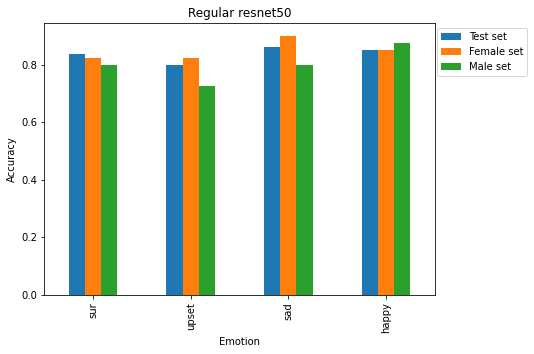}  & 
       \includegraphics[width=0.23\textwidth]{ 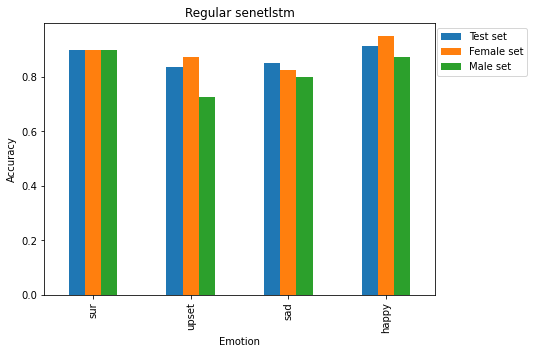}\\
       \includegraphics[width=0.23\textwidth]{ 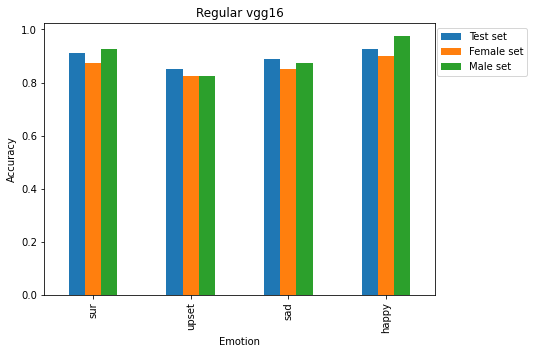}  & 
       \includegraphics[width=0.23\textwidth]{ 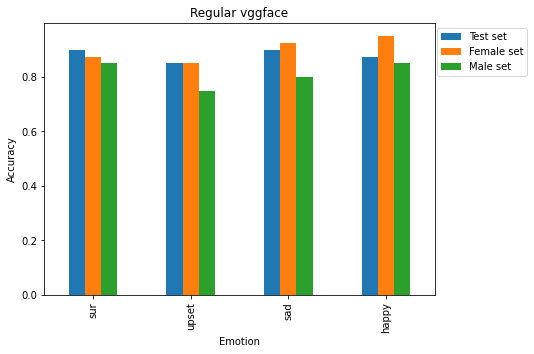}\\
    \end{tabular}
    \caption{Per model architecture, accuracy comparison between all three test sets. These are Regular models, which have been trained on entire train data.}
    \label{fig:fixed_model_regular}
\end{figure}

From the perspective of emotions (Figure~\ref{fig:fixed_model_regular}), results as follows: classification of Surprise is better for males, Upset and Sad are expressed better by females and Happy is almost identical recognized for both genders. 

\begin{figure}[!h]
    \centering
    \begin{tabular}{c c c}

       \includegraphics[width=0.14\textwidth]{ 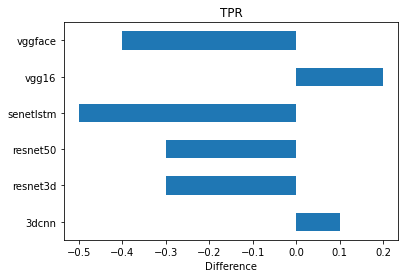} &
       \includegraphics[width=0.14\textwidth]{ 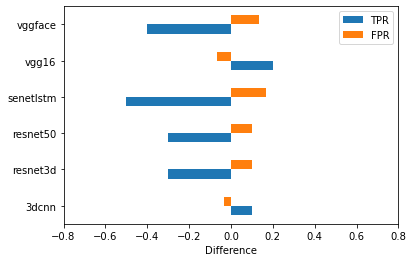} &
       \includegraphics[width=0.14\textwidth]{ 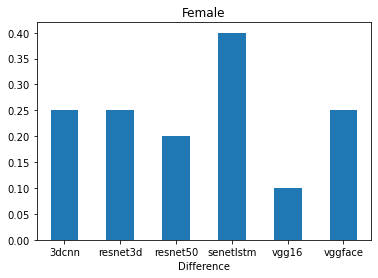}\\
       (a) & (b) & (c)\\
    \end{tabular}
    \caption{Metrics for models, which have been trained on only female data}
    \label{fig:metrics_female}
\end{figure}

Trained only on female data, Female models show completely different picture. For all three definition of fairness, \textit{SENetLSTM} architecture is considered as the most gender biased. \textit{3DCNN} is the least biased according to EQOP and EQOD (Figure~\ref{fig:metrics_female} (a,b)), and \textit{VGG16} according to the DP (Figure~\ref{fig:metrics_female} (c)). For each test set, classification accuracy is the worst for Upset emotion, Happy is the best recognised for Test set and Female set. Inference on Male set shows the best recognition of Surprise.

\begin{figure}
    \centering
    \includegraphics[width=0.48\textwidth]{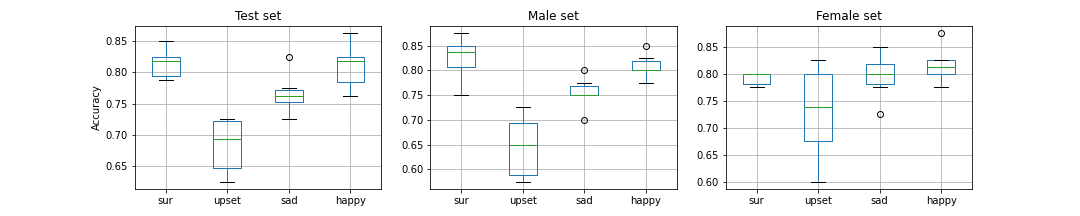}
    \caption{Aggregated accuracy for models, trained on only female data. Comparison for three different test sets.}
    \label{fig:boxplot_female}
\end{figure}

According to aggregated results (Figure~\ref{fig:boxplot_female}), variances for Male and Female set are relatively equal, therefore recognition for both gender is considered as robust. The overall accuracy is better for Male set, which is unexpected results, since models are trained exclusively on the female data.

\begin{figure}[!h]
    \centering
    \begin{tabular}{c c}
       \includegraphics[width=0.23\textwidth]{ 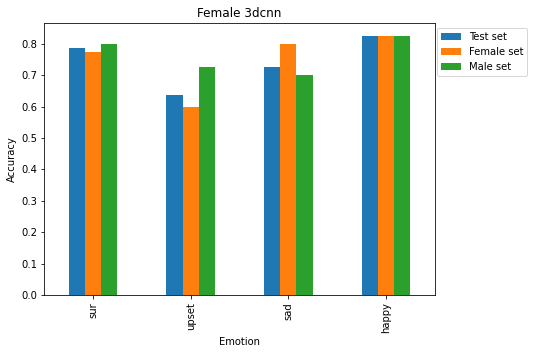}  & 
       \includegraphics[width=0.23\textwidth]{ 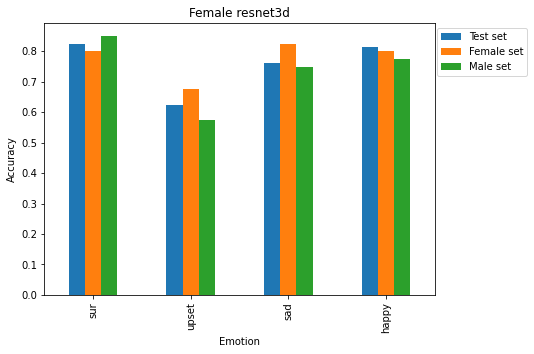}\\
       \includegraphics[width=0.23\textwidth]{ 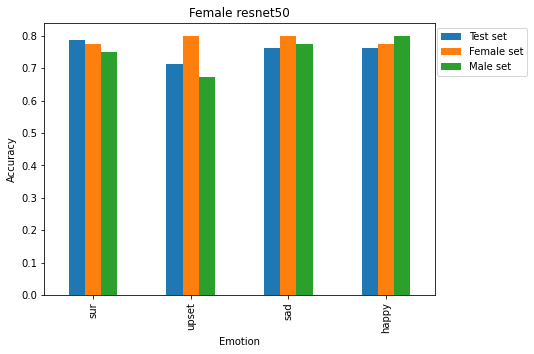}  & 
       \includegraphics[width=0.23\textwidth]{ 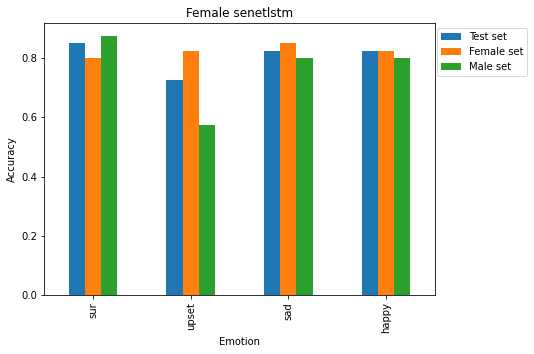}\\
       \includegraphics[width=0.23\textwidth]{ 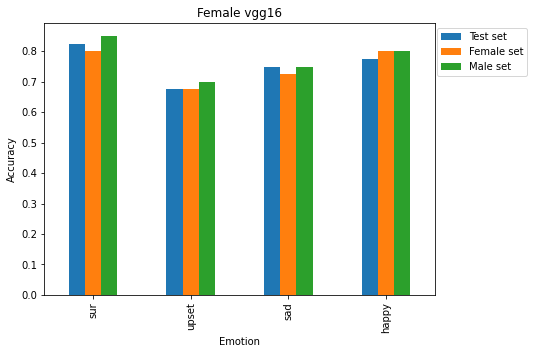}  & 
       \includegraphics[width=0.23\textwidth]{ 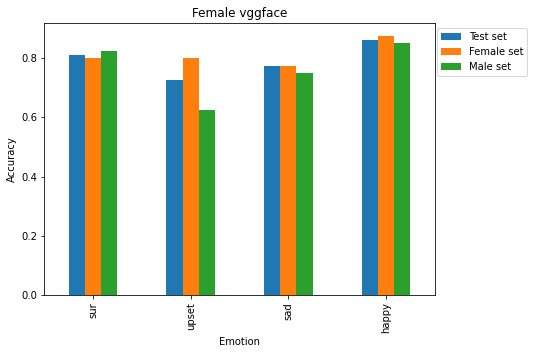}\\
    \end{tabular}
    \caption{Per model architecture, accuracy comparison between all three test sets. These are Female models, which have been trained on only female data.}
    \label{fig:fixed_model_female}
\end{figure}

In the per emotion competition (Figure~\ref{fig:fixed_model_female}), classification of Happy and Surprise emotions are almost identical for both genders, meanwhile Upset and Sad are better recognized on female subjects.

\begin{figure}[!h]
    \centering
    \begin{tabular}{c c c}

       \includegraphics[width=0.15\textwidth]{ 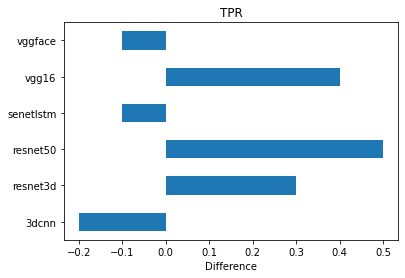} &
       \includegraphics[width=0.15\textwidth]{ 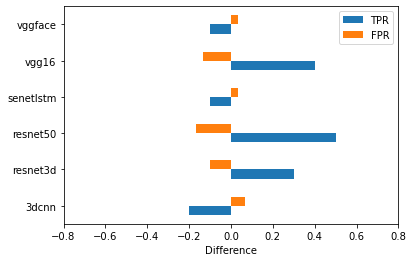} &
       \includegraphics[width=0.15\textwidth]{ 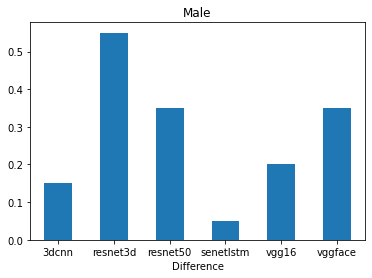}\\
       (a) & (b) & (c)\\
    \end{tabular}
    \caption{Metrics for models, which have been trained on only male data}
    \label{fig:metrics_male}
\end{figure}

Opposite to previously mentioned type of models, Male models are trained on male data. \textit{SENetLSTM} architecture is shown as the most fair architecture, according to the EQOP, EQOD and DP. For the most biased model, \textit{ResNet50} is considered of being so with respect to EQOP and EQOD, while \textit{ResNet3D} is a choice according to  DP (Figure~\ref{fig:metrics_male}). Noticeable, the according to DP, \textit{ResNet50} and \textit{VGGFACE} share second the most biased place in ranking. Therefore, results for Male models are considered to be consistent. For different test sets, the best and the worst accurate classified emotion are the same. Upset emotion has the lowest accuracy among all test sets and Happy - the highest one.

\begin{figure}[!h]
    \centering
    \includegraphics[width=0.48\textwidth]{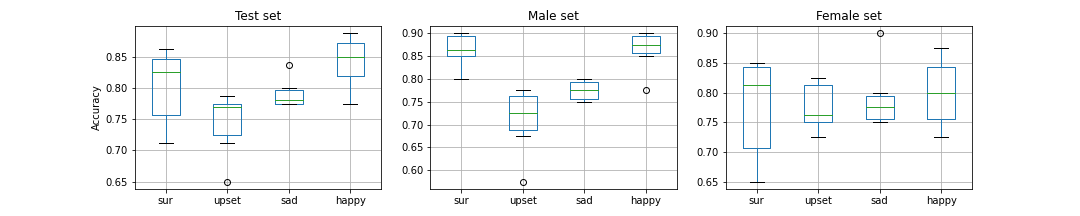}
    \caption{Aggregated accuracy for models, trained on only male data. Comparison for three different test sets.}
    \label{fig:boxplot_male}
\end{figure}

Aggregate results (Figure~\ref{fig:boxplot_male}) show that inference on Female set has high variance, being unstable, while inference on Male set has small variance, and therefore, more robust. Without a surprise here, average accuracy on Male set is higher rather than on Female set.

\begin{figure}[!h]
    \centering
    \begin{tabular}{c c}
       \includegraphics[width=0.23\textwidth]{ 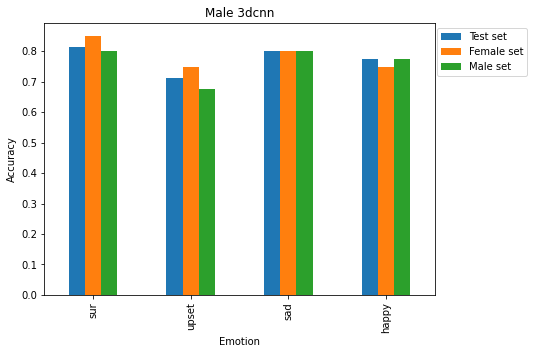}  & 
       \includegraphics[width=0.23\textwidth]{ 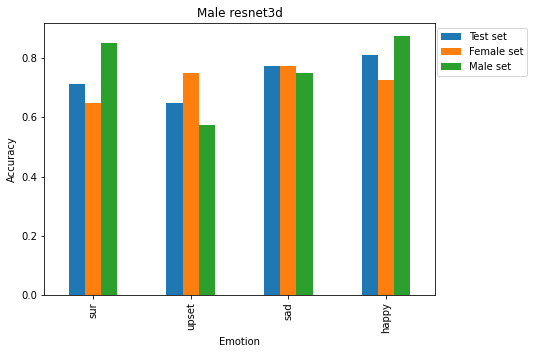}\\
       \includegraphics[width=0.23\textwidth]{ 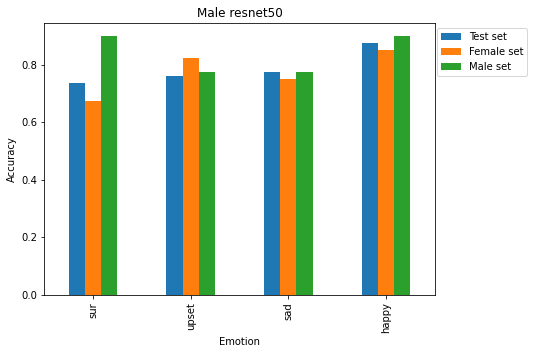}  & 
       \includegraphics[width=0.23\textwidth]{ 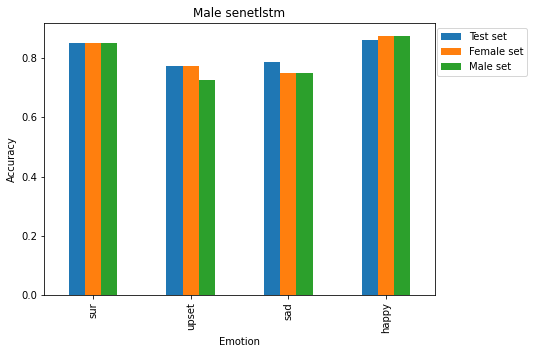}\\
       \includegraphics[width=0.23\textwidth]{ 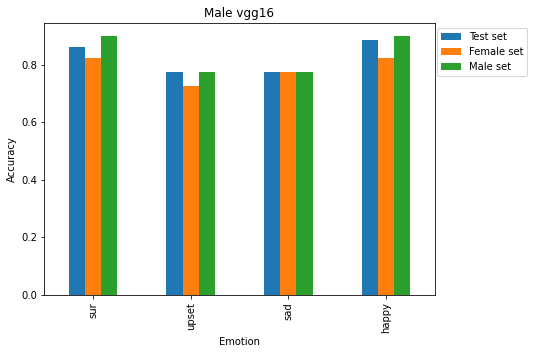}  & 
       \includegraphics[width=0.23\textwidth]{ 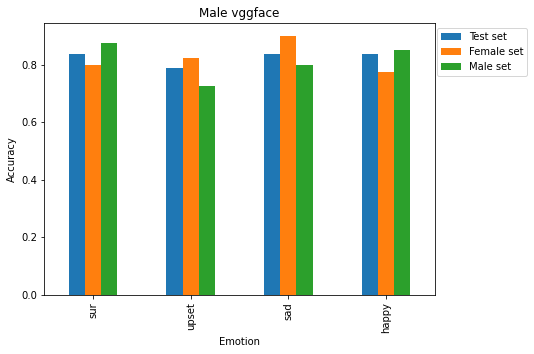}\\
    \end{tabular}
    \caption{Per model architecture, accuracy comparison between all three test sets. These are Male models, which have been trained on only male data.}
    \label{fig:fixed_model_male}
\end{figure}

In the per emotion accuracy (Figure~\ref{fig:fixed_model_male}), inference on Male set leads in recognition of Surprise and Happy emotions. Sad recognition rate is almost the same for both sets. Accuracy for Upset emotion is much higher for Female set.

\section{Conclusion}
This paper provided a comprehensive overview on the face emotion recognition biases in the context of reliable AI. Taking the concrete dataset, SASE-FE, two different groups of methods for face emotion recognition have been analyzed on gender bias. Each group consists of three different neural network architectures, where some of them have been ready available and some have been manually implemented before the analysis. The test sets have been organized in three different ways: entire test data, only male data and only female data. The train sets have been organized in the similar way. All architectures have been trained, resulting into 18 different models, 6 architectures per each train set. 

Since model bias can be explained through fairness, there have been given three different definition of fairness, according to which proper analysis have been conducted. As a results, it has been found which architectures are most biased with respect to the definitions of fairness, and which ones are more likely to be fair. In addition, it has been discovered, which kinds of emotions are easier to recognize for men and women. In addition, using three distinct train sets the relationship between training data and inference has been analyzed.

The topic of gender bias is relatively young and not addressed properly. Therefore, the amount of existed probable directions to research is immense. Extending this research work, for sure, the next goal has to be to expand research on other databases, which are widely used in FER. Also, the models which have been utilized in this paper, are not competitors to the state-of-the-art solutions. Hence, another direction of future work is to implement these solutions and analyze whether they comprise gender bias. Far-reaching extensions include other aspects of RAI and XAI. For example study of other biases (race, age, culture) or working on explanation, understanding and discovering knowledge limits. Altogether, these researches will create a background to more standardized regulation and law creation in the field of AI. As a result, integration of AI in society will be reliable and safe. After all, modern AI still encompasses a decent amount of unknown and hazard, therefore future us have to be ready.

\section*{Acknowledgements}
This work is supported by the Estonian Centre of Excellence in IT (EXCITE) funded by the European Regional Development Fund. The authors also gratefully acknowledge the support of NVIDIA Corporation with the donation of the Titan XP Pascal GPU.

\printbibliography

\vskip -2\baselineskip plus -1fil
\begin{IEEEbiography}[{\includegraphics[width=0.8in,height=0.9in,clip]{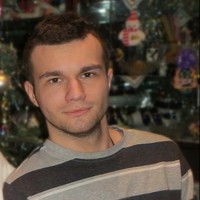}}]{Artem Domnich} is MSc in Computer Science, received degree from University of Tartu, Estonia. He is currently working in Microsoft Inc. He has several year of industrial experience in Computer Vision. In 2019, he completed internship in IBM Belgium, followed by exchange study in KTH, Sweden. During 2020, he has been working as a programmer in University of Tartu, developing solutions for autonomous vehicles. His research works involves deep learning neural networks for image and video processing, such as end-to-end driving, human detection on thermal images and face recognition. 
\end{IEEEbiography}
\vskip -2\baselineskip plus -1fil
\begin{IEEEbiography}[{\includegraphics[width=0.9in,height=1.1in,clip]{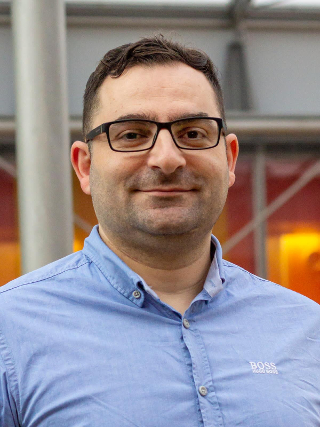}}]{Gholamreza Anbarjafari} is founder of the intelligent computer vision (iCV) lab at the University of Tartu and co-founder of iVCV. He is a Director and Chief Data Scientist at PwC Finland where he is focusing on implementation of Responsible AI in various industrial venues. He was also Deputy Scientific Coordinator of he European Network on Integrating Vision and Language (iV\&L Net) ICT COST Action IC1307. He has been a representative of Estonia for several other COST Actions. He is Associate Editor and Guest Lead Editor of several journals, Special Issues and Book projects. He is an IEEE Senior member and was the Chair of Signal Processing/Circuits and Systems/Solid-State Circuits Joint Societies Chapter of IEEE Estonian section. His research has been funded by Estonian Research Council and EU (H2020) grants and his lab has been involved in many international industrial projects. He is expert in affective computing, responsible and explainable AI, computer vision, machine learning, human-robot interaction, and artificial intelligence. He has supervised 17 MSc students and 10 PhD students. He has published over 130 scientific works. He is associate editor of several journals including Springer SIVP, MDPI Entropy, Information, and JAD. He has been TCP of conferences such as ICOSST, ICGIP, CVPR, SampTA and SIU. He has been organising challenges and workshops in FG17, CVPR17, ICCV17, ECML19, and FG20.
\end{IEEEbiography}

\newpage
\section{Appendix}
\label{appendix:tables}

\subsection{Test set}
\begin{table}[htb]
    \centering
    \caption{Regular models, Test set, accuracy}
    \begin{tabular}{|c|c|c|c|c|}
    \hline
        \textbf{Accuracy} &    Surprised  &  Upset & Sad  & Happy\\
    \hline
3dcnn  &0.8250  &0.7375  &0.8125  &0.8500\\
resnet3d  &0.8625  &0.7125  &0.8000  &0.8250\\
resnet50  &0.8375  &0.8000  &0.8625  &0.8500\\
senetlstm  &0.9000  &0.8375  &0.8500  &0.9125\\
vgg16  &0.9125  &0.8500  &0.8875  &0.9250\\
vggface  &0.9000  &0.8500  &0.9000  &0.8750\\
    \hline
    \end{tabular}
    \label{tab:rt_acc}
\end{table}

\begin{table}[h]
    \centering
    \caption{Regular models, Test set, true positive rate}
    \begin{tabular}{|c|c|c|c|c|}
    \hline
        \textbf{TPR} &    Surprised  &  Upset & Sad  & Happy\\
    \hline
        3dcnn  &0.45  &0.95  &0.30  &0.75\\
        resnet3d  &0.55  &0.70  &0.50  &0.65\\
        resnet50  &0.55  &0.85  &0.45  &0.85\\
        senetlstm  &0.70  &0.95  &0.55  &0.80\\
        vgg16  &0.75  &0.90  &0.60  &0.90\\
        vggface  &0.70  &0.80  &0.70  &0.85\\
    \hline
    \end{tabular}
    \label{tab:rt_tpr}
\end{table}

\begin{table}[h]
    \centering
    \caption{Regular models, Test set, false positive rate}
    \begin{tabular}{|c|c|c|c|c|}
    \hline
        \textbf{FPR} &    Surprised  &  Upset & Sad  & Happy\\
    \hline
3dcnn  &0.05  &0.33  &0.016  &0.12\\
resnet3d  &0.033  &0.28  &0.1 &0.12\\
resnet50  &0.066  &0.21  &0.0  &0.15\\
senetlstm  &0.033  &0.1 &0.05  &0.05\\
vgg16  &0.033  &0.16  &0.016  &0.066\\
vggface  &0.033  &0.13  &0.033  &0.12\\
    \hline
    \end{tabular}
    \label{tab:rt_fpr}
\end{table}


\begin{table}[H]
    \caption{Female models, Test set, accuracy}
    \label{tab:ft_acc}
    \centering
    \begin{tabular}{|c|c|c|c|c|}
    \hline
        \textbf{Accuracy} &    Surprised  &  Upset & Sad  & Happy\\
    \hline
3dcnn &0.7875 &0.6375 &0.7250 &0.8250\\
resnet3d &0.8250 &0.6250 &0.7625 &0.8125\\
resnet50 &0.7875 &0.7125 &0.7625 &0.7625\\
senetlstm &0.8500 &0.7250 &0.8250 &0.8250\\
vgg16 &0.8250 &0.6750 &0.7500 &0.7750\\
vggface &0.8125 &0.7250 &0.7750 &0.8625\\
    \hline
    \end{tabular}
\end{table}

\begin{table}[H]
    \caption{Female models, Test set, true positive rate}
    \label{tab:ft_tpr}
    \centering
    \begin{tabular}{|c|c|c|c|c|}
    \hline
        \textbf{TPR} &    Surprised  &  Upset & Sad  & Happy\\
    \hline
        3dcnn &0.35 &0.75 &0.30 &0.55\\
        resnet3d &0.40 &0.75 &0.15 &0.75\\
        resnet50 &0.25 &0.35 &0.65 &0.80\\
        senetlstm &0.45 &0.95 &0.45 &0.60\\
        vgg16 &0.35 &0.70 &0.10 &0.90\\
        vggface &0.45 &0.80 &0.40 &0.70\\
    \hline
    \end{tabular}
\end{table}

\begin{table}[H]
    \caption{Female models, test set, false positive rate}
    \label{tab:ft_fpr}
    \centering
    \begin{tabular}{|c|c|c|c|c|}
    \hline
        \textbf{FPR} &    Surprised  &  Upset & Sad  & Happy\\
    \hline
3dcnn &0.066 &0.4 &0.13 &0.08\\
resnet3d &0.033 &0.41 &0.033 &0.16\\
resnet50 &0.033 &0.16 &0.2 &0.25\\
senetlstm &0.016 &0.35 &0.05 &0.1\\
vgg16 &0.016 &0.33 &0.033 &0.26\\
vggface &0.066 &0.3 &0.1&0.08\\
    \hline
    \end{tabular}
\end{table}


\begin{table}[H]
    \centering
    \caption{Male models, Test set, accuracy}
    \begin{tabular}{|c|c|c|c|c|}
    \hline
        \textbf{Accuracy} &    Surprised  &  Upset & Sad  & Happy\\
    \hline
3dcnn &0.8125 &0.7125 &0.8000 &0.7750\\
resnet3d &0.7125 &0.6500 &0.7750 &0.8125\\
resnet50 &0.7375 &0.7625 &0.7750 &0.8750\\
senetlstm &0.8500 &0.7750 &0.7875 &0.8625\\
vgg16 &0.8625 &0.7750 &0.7750 &0.8875\\
vggface &0.8375 &0.7875 &0.8375 &0.8375\\
    \hline
    \end{tabular}

    \label{tab:mt_acc}
\end{table}

\begin{table}[H]
    \centering
    \caption{Male models, Test set, true positive rate}
    \begin{tabular}{|c|c|c|c|c|}
    \hline
        \textbf{TPR} &    Surprised  &  Upset & Sad  & Happy\\
    \hline
3dcnn &0.60 &0.80 &0.20 &0.60\\
resnet3d &0.55 &0.50 &0.10 &0.75\\
resnet50 &0.65 &0.60 &0.20 &0.85\\
senetlstm &0.60 &0.70 &0.45 &0.80\\
vgg16 &0.65 &0.90 &0.10 &0.95\\
vggface &0.55 &0.65 &0.55 &0.85\\
    \hline
    \end{tabular}
    
    \label{tab:mt_tpr}
\end{table}
\begin{table}[H]
    \centering
    \caption{Male models, Test set, false positive rate}
    \begin{tabular}{|c|c|c|c|c|}
    \hline
        \textbf{FPR} &    Surprised  &  Upset & Sad  & Happy\\
    \hline
3dcnn &0.12 &0.316667 &0.0 &0.16\\
resnet3d &0.23 &0.3 &0.0 &0.16\\
resnet50 &0.23 &0.183 &0.033 &0.12\\
senetlstm &0.066 &0.2 &0.1&0.12\\
vgg16 &0.066 &0.26 &0.0 &0.13\\
vggface &0.066 &0.16 &0.066 &0.16\\
    \hline
    \end{tabular}
    
    \label{tab:mt_fpr}
\end{table}

\subsection{Female set}
\begin{table}[H]
    \centering
    \caption{Regular models, Female set, accuracy}
    \begin{tabular}{|c|c|c|c|c|}
    \hline
        \textbf{Accuracy} &    Surprised  &  Upset & Sad  & Happy\\
    \hline
3dcnn  &0.875  &0.825  &0.850  &0.850\\
resnet3d  &0.825  &0.750  &0.800  &0.825\\
resnet50  &0.825  &0.825  &0.900  &0.850\\
senetlstm  &0.900  &0.875  &0.825  &0.950\\
vgg16  &0.875  &0.825  &0.850  &0.900\\
vggface  &0.875  &0.850  &0.925  &0.950\\
    \hline
    \end{tabular}
    
    \label{tab:rf_acc}
\end{table}

\begin{table}[H]
    \centering
    \caption{Regular models, Female set, true positive rate}
    \begin{tabular}{|c|c|c|c|c|}
    \hline
        \textbf{TPR} &    Surprised  &  Upset & Sad  & Happy\\
    \hline
3dcnn  &0.6  &1.0  &0.4  &0.8\\
resnet3d  &0.4  &0.5  &0.7  &0.8\\
resnet50  &0.4  &0.8  &0.7  &0.9\\
senetlstm  &0.6  &0.9  &0.6  &1.0\\
vgg16  &0.7  &0.7  &0.6  &0.9\\
vggface  &0.8  &0.4  &1.0  &1.0\\
    \hline
    \end{tabular}
    
    \label{tab:rf_tpr}
\end{table}
\begin{table}[H]
    \centering
    \caption{Regular models, Female set, false positive rate}
    \begin{tabular}{|c|c|c|c|c|}
    \hline
        \textbf{FPR} &    Surprised  &  Upset & Sad  & Happy\\
    \hline
3dcnn  &0.033  &0.23  &0.0  &0.13\\
resnet3d  &0.033  &0.16  &0.16  &0.16\\
resnet50  &0.033  &0.16  &0.033  &0.16\\
senetlstm  &0.0  &0.13  &0.1 &0.066\\
vgg16  &0.066  &0.13  &0.066  &0.1\\
vggface  &0.1 &0.0  &0.1 &0.066\\
    \hline
    \end{tabular}
    
    \label{tab:rf_fpr}
\end{table}


\begin{table}[H]
    \centering
    \caption{Regular models, Female set, accuracy}
    \begin{tabular}{|c|c|c|c|c|}
    \hline
        \textbf{Accuracy} &    Surprised  &  Upset & Sad  & Happy\\
    \hline
3dcnn &0.775 &0.600 &0.800 &0.825\\
resnet3d &0.800 &0.675 &0.825 &0.800\\
resnet50 &0.775 &0.800 &0.800 &0.775\\
senetlstm &0.800 &0.825 &0.850 &0.825\\
vgg16 &0.800 &0.675 &0.725 &0.800\\
vggface &0.800 &0.800 &0.775 &0.875\\
    \hline
    \end{tabular}
    
    \label{tab:ff_acc}
\end{table}

\begin{table}[H]
    \centering
    \caption{Regular models, Female set, true positive rate}
    \begin{tabular}{|c|c|c|c|c|}
    \hline
        \textbf{TPR} &    Surprised  &  Upset & Sad  & Happy\\
    \hline
3dcnn &0.4 &1.0 &0.2 &0.4\\
resnet3d &0.2 &0.7 &0.3 &1.0\\
resnet50 &0.1 &0.5 &0.7 &1.0\\
senetlstm &0.3 &0.7 &0.7 &0.9\\
vgg16 &0.2 &0.7 &0.2 &0.9\\
vggface &0.3 &0.5 &0.7 &1.0\\
    \hline
    \end{tabular}
    
    \label{tab:ff_tpr}
\end{table}
\begin{table}[H]
    \centering
    \caption{Regular models, Female set, false positive rate}
    \begin{tabular}{|c|c|c|c|c|}
    \hline
        \textbf{FPR} &    Surprised  &  Upset & Sad  & Happy\\
    \hline
3dcnn &0.1&0.533333 &0.0 &0.033\\
resnet3d &0.0 &0.33 &0.0 &0.26\\
resnet50 &0.0 &0.1&0.16 &0.3\\
senetlstm &0.033 &0.13 &0.1&0.2\\
vgg16 &0.0 &0.33 &0.1&0.23\\
vggface &0.033 &0.1&0.2 &0.16\\
    \hline
    \end{tabular}
    
    \label{tab:ff_fpr}
\end{table}


\begin{table}[H]
    \centering
    \caption{Regular models, Female set, accuracy}
    \begin{tabular}{|c|c|c|c|c|}
    \hline
        \textbf{Accuracy} &    Surprised  &  Upset & Sad  & Happy\\
    \hline
3dcnn &0.850 &0.750 &0.800 &0.750\\
resnet3d &0.650 &0.750 &0.775 &0.725\\
resnet50 &0.675 &0.825 &0.750 &0.850\\
senetlstm &0.850 &0.775 &0.750 &0.875\\
vgg16 &0.825 &0.725 &0.775 &0.825\\
vggface &0.800 &0.825 &0.900 &0.775\\
    \hline
    \end{tabular}
    
    \label{tab:mf_acc}
\end{table}

\begin{table}[H]
    \centering
    \caption{Regular models, Female set, true positive rate}
    \begin{tabular}{|c|c|c|c|c|}
    \hline
        \textbf{TPR} &    Surprised  &  Upset & Sad  & Happy\\
    \hline
3dcnn &0.5 &0.8 &0.2 &0.8\\
resnet3d &0.4 &0.5 &0.1 &0.8\\
resnet50 &0.6 &0.8 &0.0 &0.8\\
senetlstm &0.5 &0.5 &0.5 &1.0\\
vgg16 &0.5 &0.7 &0.1 &1.0\\
vggface &0.4 &0.4 &0.8 &1.0\\
    \hline
    \end{tabular}
    
    \label{tab:mf_tpr}
\end{table}
\begin{table}[H]
    \centering
    \caption{Regular models, Female set, false positive rate}
    \begin{tabular}{|c|c|c|c|c|}
    \hline
        \textbf{FPR} &    Surprised  &  Upset & Sad  & Happy\\
    \hline
3dcnn &0.033 &0.26 &0.0 &0.26\\
resnet3d &0.26 &0.16 &0.0 &0.3\\
resnet50 &0.3 &0.16 &0.0 &0.13\\
senetlstm &0.033 &0.13 &0.16 &0.16\\
vgg16 &0.066 &0.26 &0.0 &0.23\\
vggface &0.066 &0.033 &0.066 &0.3\\
    \hline
    \end{tabular}
    
    \label{tab:mf_fpr}
\end{table}

\subsection{Male set}

\begin{table}[H]
    \centering
    \caption{Regular models, Male set, accuracy}
    \begin{tabular}{|c|c|c|c|c|}
    \hline
        \textbf{Accuracy} &    Surprised  &  Upset & Sad  & Happy\\
    \hline
3dcnn  &0.775  &0.675  &0.775  &0.875\\
resnet3d  &0.875  &0.650  &0.825  &0.800\\
resnet50  &0.800  &0.725  &0.800  &0.875\\
senetlstm  &0.900  &0.725  &0.800  &0.875\\
vgg16  &0.925  &0.825  &0.875  &0.975\\
vggface  &0.850  &0.750  &0.800  &0.850\\
    \hline
    \end{tabular}
    
    \label{tab:rm_acc}
\end{table}

\begin{table}[H]
    \centering
    \caption{Regular models, Male set, true positive rate}
    \begin{tabular}{|c|c|c|c|c|}
    \hline
        \textbf{TPR} &    Surprised  &  Upset & Sad  & Happy\\
    \hline
3dcnn  &0.3  &1.0  &0.2  &0.7\\
resnet3d  &0.7  &0.8  &0.3  &0.5\\
resnet50  &0.7  &0.7  &0.2  &0.8\\
senetlstm  &0.8  &0.7  &0.5  &0.6\\
vgg16  &0.8  &0.9  &0.6  &0.9\\
vggface  &0.6  &0.8  &0.4  &0.7\\
    \hline
    \end{tabular}
    
    \label{tab:rm_tpr}
\end{table}
\begin{table}[H]
    \centering
    \caption{Regular models, Male set, false positive rate}
    \begin{tabular}{|c|c|c|c|c|}
    \hline
        \textbf{FPR} &    Surprised  &  Upset & Sad  & Happy\\
    \hline
3dcnn  &0.066  &0.43  &0.033  &0.066\\
resnet3d  &0.066  &0.4  &0.0  &0.1\\
resnet50  &0.16  &0.26  &0.0  &0.1\\
senetlstm  &0.066  &0.26  &0.1 &0.033\\
vgg16  &0.033  &0.2  &0.033  &0.0\\
vggface  &0.066  &0.26  &0.066  &0.1\\
    \hline
    \end{tabular}
    
    \label{tab:rm_fpr}
\end{table}


\begin{table}[H]
    \centering
    \caption{Regular models, Male set, accuracy}
    \begin{tabular}{|c|c|c|c|c|}
    \hline
        \textbf{Accuracy} &    Surprised  &  Upset & Sad  & Happy\\
    \hline
3dcnn &0.800 &0.725 &0.700 &0.825\\
resnet3d &0.850 &0.575 &0.750 &0.775\\
resnet50 &0.750 &0.675 &0.775 &0.800\\
senetlstm &0.875 &0.575 &0.800 &0.800\\
vgg16 &0.850 &0.700 &0.750 &0.800\\
vggface &0.825 &0.625 &0.750 &0.850\\
    \hline
    \end{tabular}
    
    \label{tab:fm_acc}
\end{table}

\begin{table}[H]
    \centering
    \caption{Regular models, Male set, true positive rate}
    \begin{tabular}{|c|c|c|c|c|}
    \hline
        \textbf{TPR} &    Surprised  &  Upset & Sad  & Happy\\
    \hline
3dcnn &0.3 &0.7 &0.4 &0.7\\
resnet3d &0.6 &0.8 &0.0 &0.5\\
resnet50 &0.4 &0.4 &0.6 &0.6\\
senetlstm &0.6 &1.0 &0.2 &0.3\\
vgg16 &0.5 &0.8 &0.0 &0.9\\
vggface &0.6 &1.0 &0.1 &0.4\\
    \hline
    \end{tabular}
    
    \label{tab:fm_tpr}
\end{table}
\begin{table}[H]
    \centering
    \caption{Regular models, Male set, false positive rate}
    \begin{tabular}{|c|c|c|c|c|}
    \hline
        \textbf{FPR} &    Surprised  &  Upset & Sad  & Happy\\
    \hline
3dcnn &0.033 &0.26 &0.2 &0.13\\
resnet3d &0.066 &0.5 &0.0 &0.13\\
resnet50 &0.13 &0.23 &0.16 &0.13\\
senetlstm &0.033 &0.56 &0.0 &0.033\\
vgg16 &0.033 &0.33 &0.0 &0.23\\
vggface &0.1&0.5 &0.033 &0.0\\
    \hline
    \end{tabular}
    
    \label{tab:fm_fpr}
\end{table}


\begin{table}[H]
    \centering
    \caption{Regular models, Male set, accuracy}
    \begin{tabular}{|c|c|c|c|c|}
    \hline
        \textbf{Accuracy} &    Surprised  &  Upset & Sad  & Happy\\
    \hline
3dcnn &0.800 &0.675 &0.800 &0.775\\
resnet3d &0.850 &0.575 &0.750 &0.875\\
resnet50 &0.900 &0.775 &0.775 &0.900\\
senetlstm &0.850 &0.725 &0.750 &0.875\\
vgg16 &0.900 &0.775 &0.775 &0.900\\
vggface &0.875 &0.725 &0.800 &0.850\\
    \hline
    \end{tabular}
    
    \label{tab:mm_acc}
\end{table}

\begin{table}[H]
    \centering
    \caption{Regular models, Male set, true positive rate}
    \begin{tabular}{|c|c|c|c|c|}
    \hline
        \textbf{TPR} &    Surprised  &  Upset & Sad  & Happy\\
    \hline
3dcnn &0.7 &0.8 &0.2 &0.4\\
resnet3d &0.7 &0.6 &0.1 &0.7\\
resnet50 &0.7 &0.7 &0.4 &0.9\\
senetlstm &0.7 &0.7 &0.4 &0.6\\
vgg16 &0.8 &0.9 &0.1 &0.9\\
vggface &0.7 &0.8 &0.3 &0.7\\
    \hline
    \end{tabular}
    
    \label{tab:mm_tpr}
\end{table}
\begin{table}[H]
    \centering
    \caption{Regular models, Male set, false positive rate}
    \begin{tabular}{|c|c|c|c|c|}
    \hline
        \textbf{FPR} &    Surprised  &  Upset & Sad  & Happy\\
    \hline
3dcnn &0.16 &0.36 &0.0 &0.1\\
resnet3d &0.1&0.43 &0.033 &0.066\\
resnet50 &0.033 &0.2 &0.1&0.1\\
senetlstm &0.1&0.26 &0.13 &0.033\\
vgg16 &0.066 &0.26 &0.0 &0.1\\
vggface &0.066 &0.3 &0.033 &0.1\\
    \hline
    \end{tabular}
    
    \label{tab:mm_fpr}
\end{table}

\subsection{Charts}
\label{appendix:charts}

\begin{figure}[!h]
    \centering
    \begin{tabular}{c c}
       \includegraphics[width=0.23\textwidth]{ 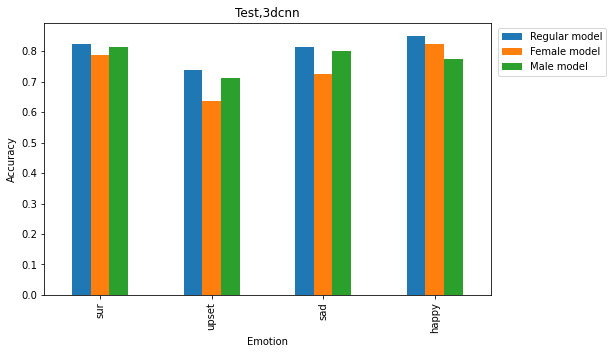}  & 
       \includegraphics[width=0.23\textwidth]{ 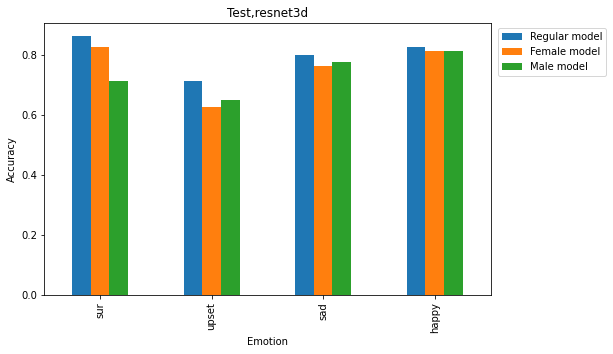}\\
       \includegraphics[width=0.23\textwidth]{ 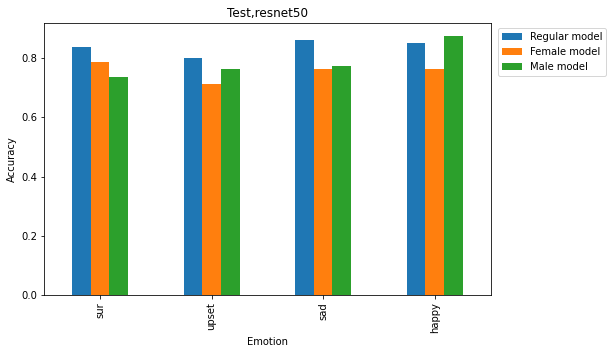}  & 
       \includegraphics[width=0.23\textwidth]{ 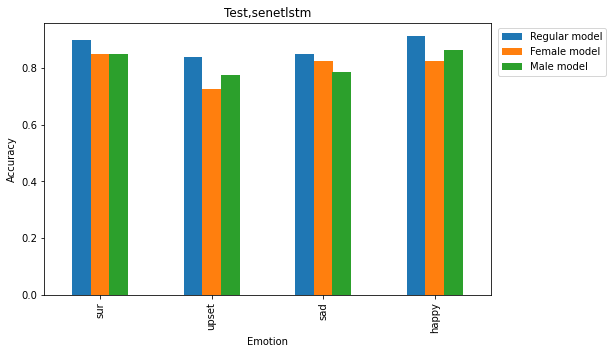}\\
       \includegraphics[width=0.23\textwidth]{ 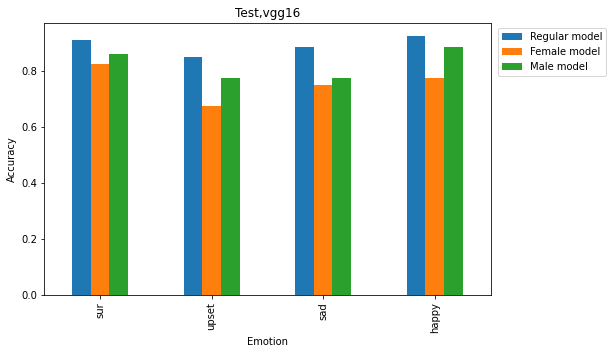}  & 
       \includegraphics[width=0.23\textwidth]{ 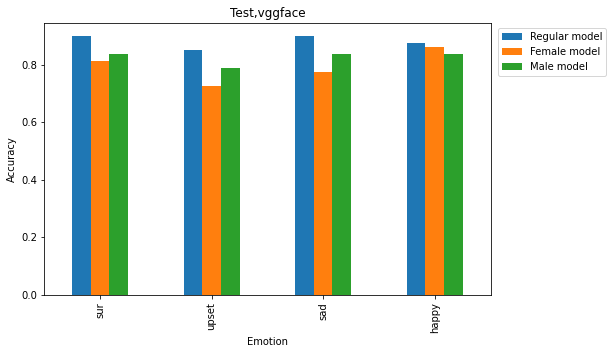}\\
    \end{tabular}
    \caption{Per model architecture, Test set accuracy comparison between different model types}
    \label{fig:fixed_set_test}
\end{figure}

\begin{figure}[!h]
    \centering
    \begin{tabular}{c c}
       \includegraphics[width=0.23\textwidth]{ 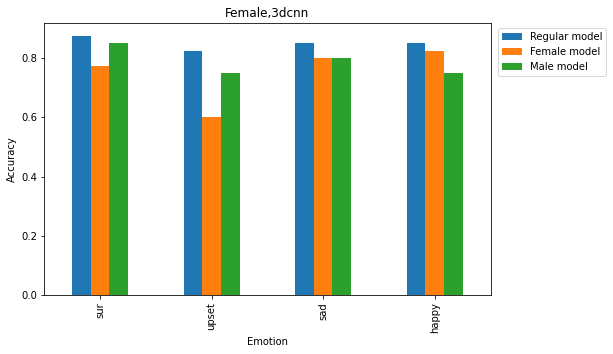}  & 
       \includegraphics[width=0.23\textwidth]{ 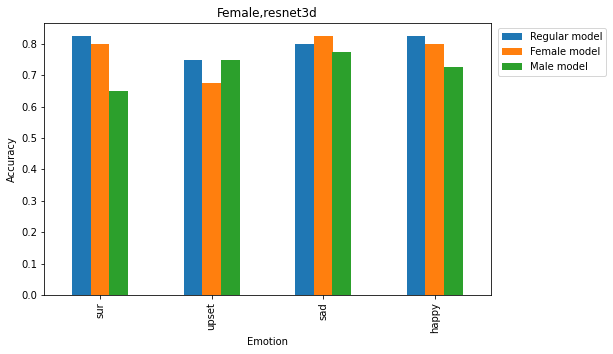}\\
       \includegraphics[width=0.23\textwidth]{ 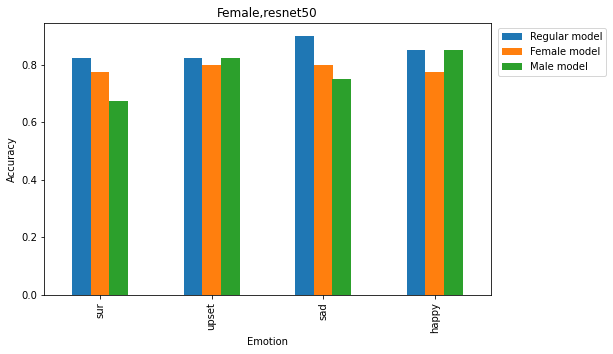}  & 
       \includegraphics[width=0.23\textwidth]{ 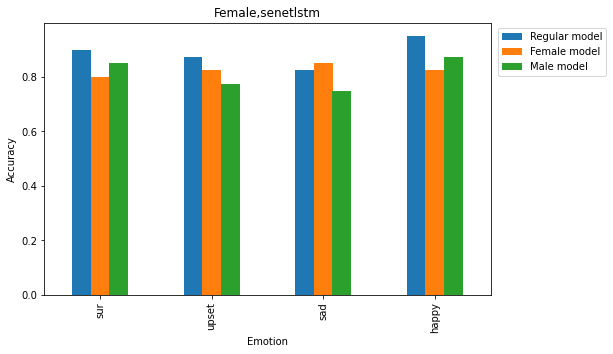}\\
       \includegraphics[width=0.23\textwidth]{ 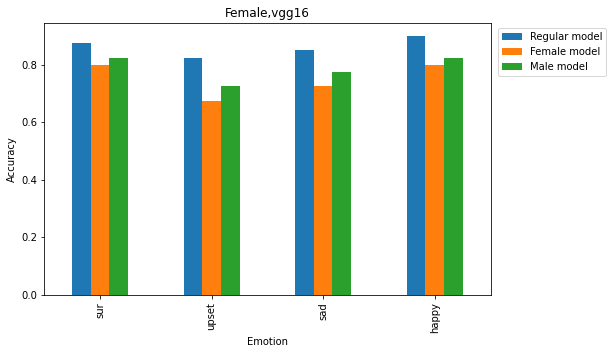}  & 
       \includegraphics[width=0.23\textwidth]{ 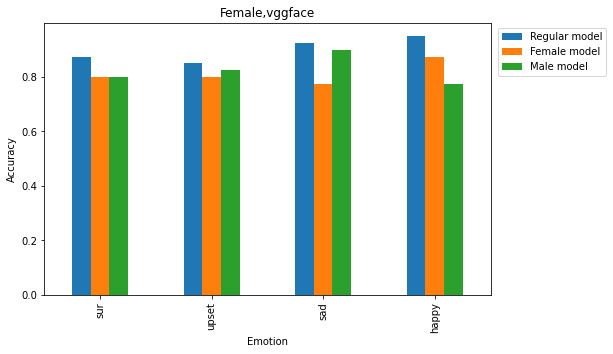}\\
    \end{tabular}
    \caption{Per model architecture, Female set accuracy comparison between different model types}
    \label{fig:fixed_set_female}
\end{figure}

\begin{figure}[!h]
    \centering
    \begin{tabular}{c c}
       \includegraphics[width=0.23\textwidth]{ 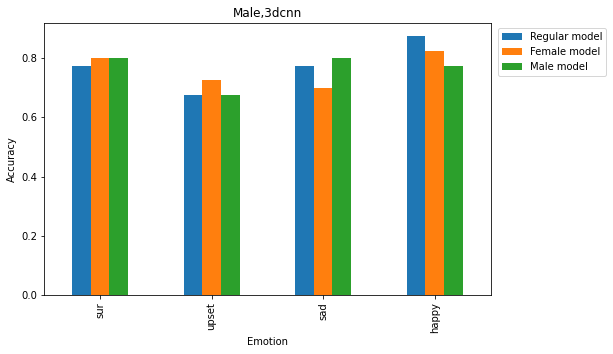}  & 
       \includegraphics[width=0.23\textwidth]{ 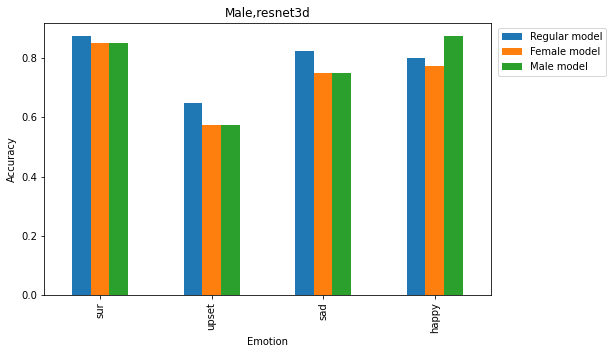}\\
       \includegraphics[width=0.23\textwidth]{ 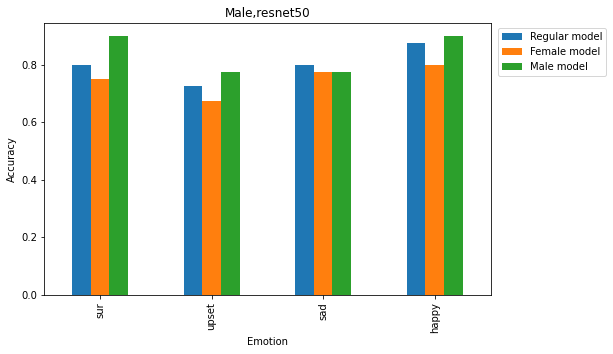}  & 
       \includegraphics[width=0.23\textwidth]{ 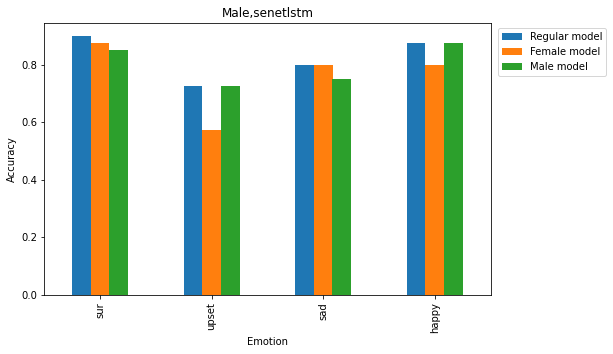}\\
       \includegraphics[width=0.23\textwidth]{ 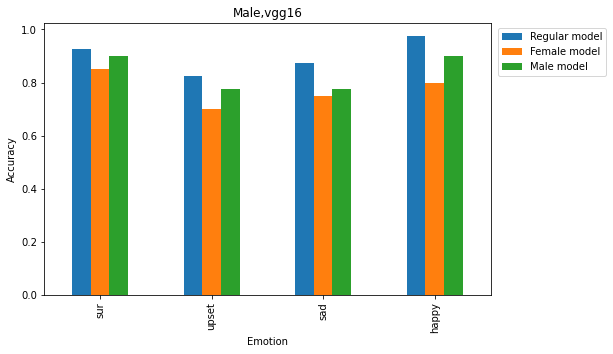}  & 
       \includegraphics[width=0.23\textwidth]{ 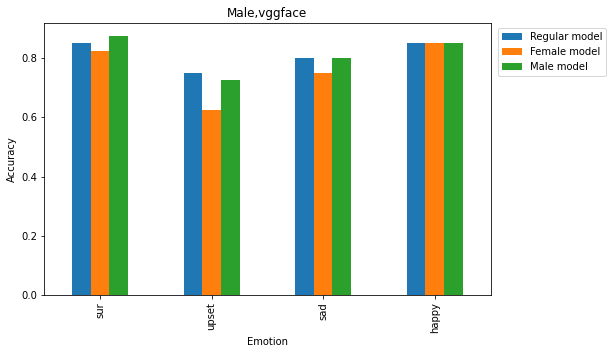}\\
    \end{tabular}
    \caption{Per model architecture, Male set accuracy comparison between different model types}
    \label{fig:fixed_set_male}
\end{figure}

\end{document}